\documentclass[10pt,journal,compsoc]{IEEEtran}
\usepackage{graphicx}
\usepackage{hyperref}

\usepackage{amsmath,amssymb}
\usepackage{bm}
\usepackage{bbold}
\usepackage{empheq}

\usepackage{multirow}
\usepackage{subcaption}

\newcommand{\norm}[1]{\left\lVert#1\right\rVert}
\newcommand\myeq{\stackrel{\mathclap{\normalfont\mbox{def}}}{=}}
\newcommand*{\vertbar}{\rule[-1ex]{0.5pt}{2.5ex}}

\renewcommand{\footnoterule}{%
  \kern -3pt
  \hrule width \linewidth height 0.5pt
  \kern 2pt
}

\ifCLASSOPTIONcompsoc
 \usepackage[nocompress]{cite}
\else
 \usepackage{cite}
\fi

\ifCLASSINFOpdf
\else
\fi

\begin{document}

\title{Investigating Pose Representations and Motion Contexts Modeling for 3D Motion Prediction}

\author{Zhenguang Liu$^*$,
Shuang Wu$^*$, Shuyuan Jin,
Shouling Ji, Qi Liu, Shijian Lu,
and Li Cheng 
\IEEEcompsocitemizethanks{
\IEEEcompsocthanksitem Zhenguang Liu is with Zhejiang Gongshang Unviersity and Zhejiang University, Hangzhou 310018, China.\protect\\ Email: liuzhenguang2008@gmail.com.
\IEEEcompsocthanksitem Shuang Wu is with the School of Computer Science and Engineering, Nanyang Technological University, Singapore 639798.\protect\\Email: wushuang@outlook.sg
\IEEEcompsocthanksitem Shuyuan Jin is with the School of Computing, National University of Singapore, Singapore 117417. Email: shuyuanjin@u.nus.edu.
\IEEEcompsocthanksitem Shouling Ji is with Zhejiang University. Email: sji@zju.edu.cn.
\IEEEcompsocthanksitem Qi Liu is with the University of Oxford. Email: qi.liu@cs.ox.ac.uk.
\IEEEcompsocthanksitem Shijian Lu is with the School of Computer Science and Engineering, Nanyang Technological University, Singapore 639798.\protect\\Email: shijian.lu@ntu.edu.sg.
\IEEEcompsocthanksitem Li Cheng is with the School of Electrical and Computer Engineering, University of Alberta. Email: lcheng5@ualberta.ca.}
\thanks{The first two authors have made equal contributions to this paper.}%
\thanks{Manuscript received 26 Oct. 2020; revised 15 Oct. 2021; accepted 27 Dec. 2021.}}

\markboth{IEEE TRANSACTIONS ON PATTERN ANALYSIS AND MACHINE INTELLIGENCE, VOL. X, NO. X, MMYYYY}
{LIU \MakeLowercase{\textit{et al.}}: Investigating Pose Representations and Motion Contexts Modeling for 3D Motion Prediction}

\IEEEtitleabstractindextext{%
\begin{abstract}
Predicting human motion from historical pose sequence is crucial for a machine to succeed in intelligent interactions with humans. One aspect that has been obviated so far, is the fact that how we represent the skeletal pose has a critical impact on the prediction results. Yet there is no effort that investigates across different pose representation schemes. We conduct an indepth study on various pose representations with a focus on their effects on the motion prediction task. Moreover, recent approaches build upon off-the-shelf RNN units for motion prediction. These approaches process input pose sequence sequentially and inherently have difficulties in capturing long-term dependencies. In this paper, we propose a novel RNN architecture termed AHMR (Attentive Hierarchical Motion Recurrent network) for motion prediction which simultaneously models local motion contexts and a global context. We further explore a geodesic loss and a forward kinematics loss for the motion prediction task, which have more geometric significance than the widely employed L2 loss. Interestingly, we applied our method to a range of articulate objects including human, fish, and mouse. Empirical results show that our approach outperforms the state-of-the-art methods in short-term prediction and achieves much enhanced long-term prediction proficiency, such as retaining natural human-like motions over 50 seconds predictions. Our codes are released.
\end{abstract}

}

\maketitle

\IEEEdisplaynontitleabstractindextext

%
\IEEEpeerreviewmaketitle

\IEEEraisesectionheading{\section{Introduction}}\label{sec:introduction}
\begin{figure*}[ht!]
\begin{center}
\includegraphics[width=0.95\textwidth]{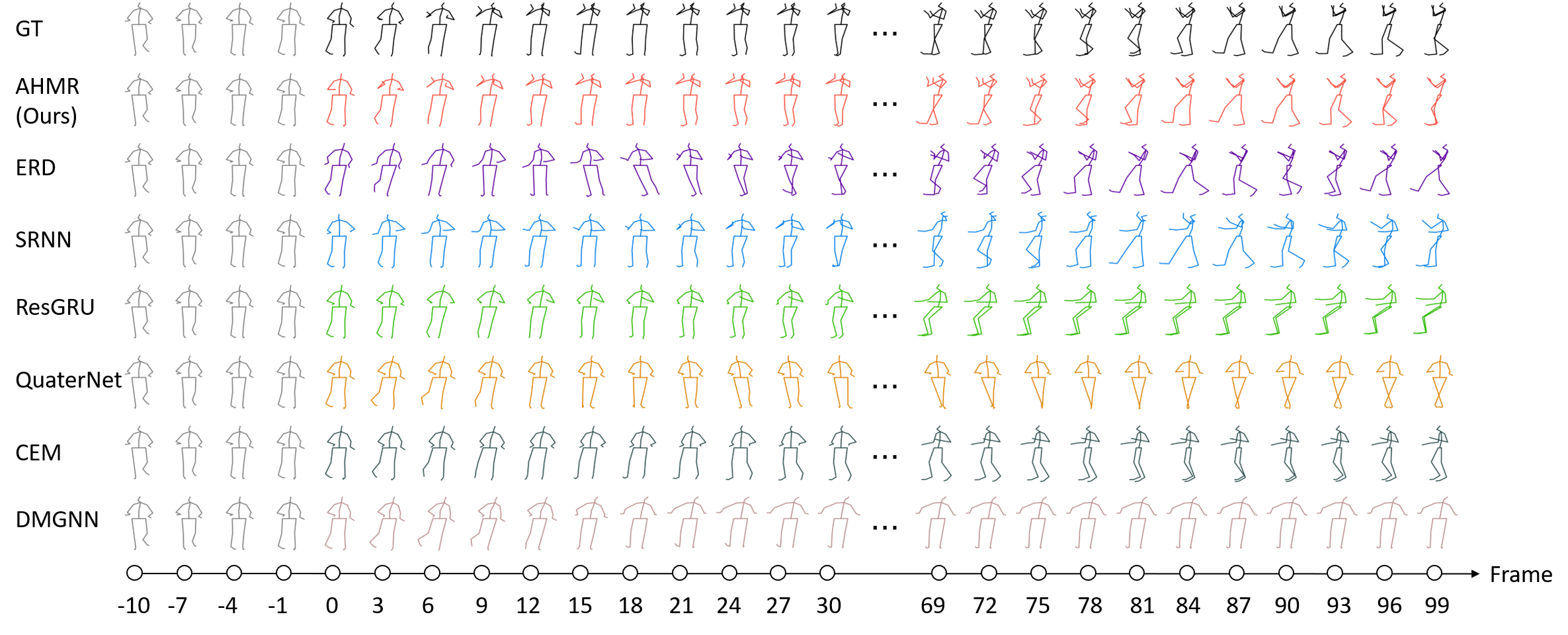}
\caption{\footnotesize{Short-term and long-term motion forecasting for eating activity on the H3.6m dataset~\cite{H36M}. Given an observed motion sequence (first 4 frames) the goal is to predict future motion (from the 5th frame). 1st line: the ground truth; 2nd line: our method; Existing methods are shown on the 3rd to 8th line: ERD~\cite{erd}, SRNN~\cite{srnn}, ResGRU~\cite{residualgru}, QuaterNet~\cite{pavllo2019modeling}, CEM~\cite{li2018convolutional} \& DMGNN~\cite{li2020dynamic}.}}
\label{fig:deficiency}
\end{center}
\end{figure*}
\IEEEPARstart{O}{ne} important component of our capacity to interact with the external world resides in our ability to predict the future~\cite{residualgru}, based on existing cues and past observations. Without this ability, it would be extremely difficult for us to hand an object to another person, avoid imminent dangers in driving, and get past defenders in sports~\cite{liu2019towards}. Likewise, anticipating the movements of surrounding objects is crucial for a machine to succeed in intelligent interactions with the physical world. Therefore, motion prediction is at the core of many applications in computer vision and robotics. For instance, predicting future motion of humans is important for autonomous driving~\cite{predictionandclassification}, where the machine cannot simply assume that a human will stay stationary at the site we see him/her. Motion prediction is also crucial for other applications such as animal tracking~\cite{erd}, assistive healthcare, and motion synthesis in games and videos~\cite{residualgru,roboticsnavigation}.

Unfortunately, unlike the motions of inanimate objects that follow deterministic physical laws, there is no simple model for the underlying conscious activities of humans or animals. The inherent complex nature of human activities~\cite{humanbehavior} invoke challenges for motion prediction in the form of \emph{non-linearity} and \emph{high dimensionality}. Over the past decade, Hidden Markov models~\cite{Markov,Boltzmann}, linear dynamic systems~\cite{Lineardynamic}, and Gaussian processes~\cite{gaussian,Gaussian2} had been popular solutions to capture motion dynamics. However, these models impose strong assumptions (\emph{e.g. Gaussian distribution assumption}) and shallow functions on motion state transition, leading to unsatisfactory motion predictions.

Recently, neural network models have become the dominant approaches in motion prediction, by virtue of their abilities to learn non-linear functions. In the framework of recurrent neural networks (RNNs),~\cite{erd} propose a Encoder-Recurrent-Decoder (ERD) network for human motion prediction. Further,~\cite{residualgru,pavllo2019modeling} employ variants of recurrent neural networks in the form of Gated Recurrent Unit (GRU) for motion prediction.~\cite{localstructure} advocates to learn local representations for different body components separately.~\cite{mao2019learning} encodes temporal contexts by working in the trajectory space instead of the traditional pose space, while~\cite{graphconvolution} exploits a dynamic graph to capture both explicit and implicit connections between joints.

Scrutinizing the released implementations of existing methods\footnote{https://gist.github.com/Seleucia/3a4f3fadc6a8dc215b4b6fd3d6c0b596;\\https://github.com/asheshjain399/RNNexp;\\ https://github.com/una-dinosauria/human-motion-prediction}, we empirically observed that their long-term predictions often degrade into motionless states or drift away to non-human-like motions, while their short-term predictions exhibit discontinuous transition from the observed pose sequence to the first frame prediction. Figure~\ref{fig:deficiency} demonstrates this phenomenon, where the first line provides the ground truth, and the 3rd to 8th lines show the motion prediction results of existing methods. We can see that, unfortunately, existing methods still suffer from severe issues including \emph{unnatural} long-term prediction and \emph{inaccurate} short-term prediction. Surprisingly, as reported in~\cite{residualgru} and presented in subsection~\ref{sec:quantitative}, quantitative evaluations even reveal that some existing methods are outperformed by a simple baseline that constantly copy the last observed pose as the future predictions. 

We conjecture that these deficiencies are mainly caused by ineffective \emph{skeletal pose representation} and \emph{motion contexts modeling}. More specifically, (1) Current approaches concentrate their attention on proposing new sorts of neural network architectures, while for pose representation they simply adopt two straightforward representations: raw 3D joint coordinates and 3D joint rotational angles. Such representations, however, are not optimal for motion prediction as they either treat the joints as independent entities or generate discontinuous embeddings that are difficult for backpropagation training. (2) Second, existing approaches heavily rely on sequential RNN units (such as LSTM and GRU)~\cite{erd,residualgru,viewadaptive}, which model motion contexts by successively reading the given pose sequence from one end to the other. These units are known to have difficulties in capturing long-term dependencies~\cite{LSTM_problem}. On another note, sequential information flow endows the architectures with non-parallel operations, which may lead to computational bottlenecks in industrial use~\cite{liuqiacl}.

To address the issue of \emph{pose representation}, we investigate across various pose representation models including 3D joint coordinates, axis angles, quaternion, and Stiefel manifold paremeterization. As will be demonstrated in our experiments, 3D joint coordinates representation is plagued by the issue of inconsistent bone lengths. In contrast, a kinematic chain representation can naturally and explicitly encode anatomical constraints such as bone length invariance and rotational degrees of freedom (DoF). Different parameterization including axis angle, quaternion, and Stiefel manifold could be utilized in the kinematic chain framework. 
By theoretical investigation and empirical study, we finally settle down to Stiefel manifold parameterization, as it provides a continuous embedding for 3D rotations that are more suitable for backpropagation in neural networks compared to other representations.

To address the issue of \emph{motion contexts modeling}, we propose a novel RNN architecture named \emph{AHMR} (\underline{A}ttentive \underline{H}ierarchical \underline{M}otion \underline{R}ecurrent Network). \emph{AHMR} does not use a single hidden state for motion contexts modeling, instead it utilizes a structured state consisting of local states for individual frames and a global state for the entire sequence. Rather than incrementally processing the input sequence as in conventional sequential RNN modules, \emph{AHMR} handles the entire input sequence concurrently and allows information exchange in multi-directions. Within each recurrent step, \emph{AHMR} updates the structured state by exchanging information between neighboring local states, and between local and global states. Significantly, spatial attention is incorporated to take into account the fact that different joints engage differently in a motion (\emph{e.g., in eating activity, the arms engage the most while other body parts usually stay motionless}). Temporal attention~\cite{Attentionisallyouneed} is also incorporated to consider temporal relationships in modeling motion dynamics.

We would like to point out that the state-of-the-art methods typically focus on \emph{human} motion prediction whereas studies on \emph{animals} are lacking. This motivates us to fill this research gap by addressing motion prediction across different object categories. In particular, we attempt to address motion prediction task on a wide range of articulated objects including fish, mouse, and human. Extensive experiments show that our approach achieves the state-of-the-art results on the large H3.6m benchmark dataset, capable of predicting natural human-like motions over a long term (50 seconds), and significantly outperforms existing methods on complex fish and mouse datasets.

To summarize, the key contributions of this paper are as follows:
(1) To the best of our knowledge, we are the first to systematically explore the limitations and strengths of different skeletal pose representation schemes on the motion prediction task. We present comprehensive theoretical analysis and interesting empirical insights, which would serve as guidance for future research in this field.
(2) We propose a novel RNN structure with non-sequential motion-context modeling and spatial temporal attention, which is able to capture rich local and global motion contexts. We also explore two theoretically motivated loss functions, a geodesic loss and a forward kinematics loss, which are better catered for optimizing the network model compared to the commonly adopted Euclidean loss.
(3) Our method sets the new state-of-the-art performance on short-term and long-term motion predictions. Our implementations and resultant videos are released at \url{https://github.com/BII-wushuang/Lie-Group-Motion-Prediction}.

\section{Related Work}
Over the years, an increasing number of approaches have been developed to address the critical problems of 3D human motion analysis. One line of works deal with video datasets~\cite{awad2018trecvid,pellegrini2010improving} such as predicting human activities~\cite{koppula2015anticipating,srivastava2015unsupervised,liang2019peeking} or forecasting pedestrian trajectories~\cite{rudenko2020human,manh2018scene}. It is beyond the scope of this paper to deliver an exhaustive report on all existing research efforts in this field. Instead, we focus on a relatively succinct account of the works centered on motion prediction from pose sequences inputs. Three dimensional pose sequences can now be conveniently acquired by commodity motion capture systems or extracted from depth images and videos using pose estimation algorithms (e.g., [24], [25], [26]). In general, the closely related work can be roughly grouped into three categories, namely
1) \emph{skeleton-based human pose representation},
2) \emph{human motion prediction}, 
3) \emph{animal motion analysis}.

\textbf{Skeleton-based Human Pose Representation} \quad Human pose representation is a fundamental problem in computer vision and robotics. Skeleton-based human pose representations have attracted intense attention due to their realtime performance~\cite{representation4} and robustness to viewpoint change, motion speed and body scale.

A pervasive solution is using raw 3D joint positions to represent human pose, as adopted in~\cite{predictionandclassification} and~\cite{representation1}.~\cite{representation5} extends this line of work by advocating to utilize only the positions of most informative joints rather than all the joints.~\cite{representation4} divides the human skeleton into five body parts comprising four limbs and the spine. Besides considering absolute 3D joint positions w.r.t. a global reference frame, relative displacement between joints has also been explored.~\cite{representation6,representation8} characterize the whole body configuration as the relative displacements between each pair of joints, whereas~\cite{representation10} considers the relative displacement of each joint w.r.t. a global reference joint (hip center). Another line of work models the skeleton as a kinematic tree and characterizes the relative orientation of bones at each joint.~\cite{representation2,srnn,GRU,mhu} parameterize the rotation of the bone with $\mathfrak{so}(3)$ Lie algebra or axis angles representations while~\cite{pavllo2019modeling} adopt a quaternion representation.~\cite{zhou2019continuity} studied yet another rotation representation formalism based on the $\mathbf{V}_2(\mathbb{R}^3)$ Stiefel manifold.

\textbf{Human Motion Prediction} \quad Traditionally, hidden Markov models~\cite{Markov,Boltzmann}, linear dynamic systems~\cite{Lineardynamic} and Gaussian processes~\cite{gaussian,Gaussian2} are introduced to capture temporal dynamics of human motions. Recently, driven by the advances of deep learning architectures and large-scale public datasets, various deep learning methods have been proposed with much superior performance~\cite{erd, srnn, residualgru, pavllo2019modeling, AGED, li2018convolutional, mao2019learning, li2020dynamic, aksan2020spatio}.
\cite{erd} presents a sequence-to-sequence model which utilizes LSTM units as encoders and decoders for motion prediction.~\cite{srnn} characterizes the human skeleton as a structured spatio-temporal graph, employing LSTM units to model body parts as nodes and spatio-temporal interactions between body parts as edges.~\cite{residualgru, pavllo2019modeling} adopt GRU for context modeling and incorporate residual connections, thereby translating the problem from predicting joint positions to predicting joint velocities. In additional to RNN based approaches,~\cite{li2018convolutional} uses convolutional networks,~\cite{aksan2020spatio} utilises the transformer network while~\cite{mao2019learning, li2020dynamic,graphconvolution} employ graph convolutional networks for generating future motion sequences.~\cite{AGED} propose incorporating adversarial training for generating more smooth predictions.

Existing works approach the motion prediction problem by modeling the motion contexts with either conventional RNN units or feedforward networks such as convolutional or graph convolutional networks. However, on a temporal level, the sequential processing of the input pose sequence in existing approaches is problematic in modeling long term dependencies. While transformer networks may handle long term dependencies, they tend to under-perform when trained on smaller datasets. In our approach, we simultaneously model local and global contexts using frame-level states and a sequence-level state in parallel. We then enrich the motion contexts incrementally by exchanging information between neighboring local states, and between local and global states. Furthermore, our approach incorporates spatial and temporal attention to model relationships between joints and frames, respectively.

\textbf{Animal Motion Analysis} \quad Now, let us consider the other two articulated objects, namely fish and mouse, to be studied in this paper. The zebrafish and lab mouse are important model organisms in the life science community and especially pertinent to the field of computational ethology~\cite{anderson2014toward}.~\cite{Mousebehavior2,Mousebehavior1} are two recent work analyzing mice social behaviors, employing a straight-line and an ellipse respectively to model the mouse skeleton.~\cite{mousefishpose,Multiplemouse} characterize fish and mouse with simplified 2D models. A few efforts conduct cross objects studies such as~\cite{Cascaded} which performs pose estimation of zebrafish, lab mouse, and human face, while~\cite{Liex} performs pose estimation on fish, mouse and human hand. However, to the best of our knowledge, there is still a lack of research on the \emph{animal motion prediction} task.

\section{Preliminaries: Skeletal Pose Parameterization Schemes} \label{sec:pose}
Broadly, a skeletal pose may be represented in two frameworks: 1) via 3D Cartesian coordinates of the keypoints; 2) via the orientation/rotation of the bones characterized along kinematic chains. In what follows, we outline their formalism and discuss their feasibility for the motion prediction task from a theoretical perspective.

\subsection{Representing Articulated Object Pose with Cartesian Coordinates}
The most intuitive and direct way to represent a skeletal pose configuration is via 3D Cartesian coordinates of the joint positions. This is typically the go-to representation employed in pose estimation tasks. Despite its simplicity, this representation is not favourable for describing motion dynamics. It regards all joints as independent entities and fails to model the constraints inherent to the skeleton, such as the bone lengths and physical restrictions at the joint rotations. Motion predictions within this representation framework lead to severe body distortion, e.g., zigzag spines and varying bone lengths in predictions, as we will illustrate in the last row of Figure~\ref{fig:walking}. Furthermore, slight errors at earlier predictions tend to propagate out of control in long-term predictions.

\subsection{Representing Articulated Object Pose with Kinematic Chains}
Moving away from the unsuitable Cartesian coordinates representation in 3D space, we formalize how to characterize the skeletal pose of an articulated object with kinematic chains. A kinematic chain consists of a chain of successive bones as well as accompanying transformations relating the bones in the chain. A remarkable feature of this approach is that it represents a skeletal chain as an organic entity, taking care of the bone length and rotational constraints, thus paving the way to the motion prediction task. We also note that the kinematic chain representation is an umbrella framework for which various parameterization schemes, \emph{such as axis angle, quaternion, and Stiefel manifold (as we will cover in the following subsections)}, may be employed to characterize the kinematic chain transformations.

\begin{figure*}[t]
\begin{center}
\includegraphics[width=1\linewidth]{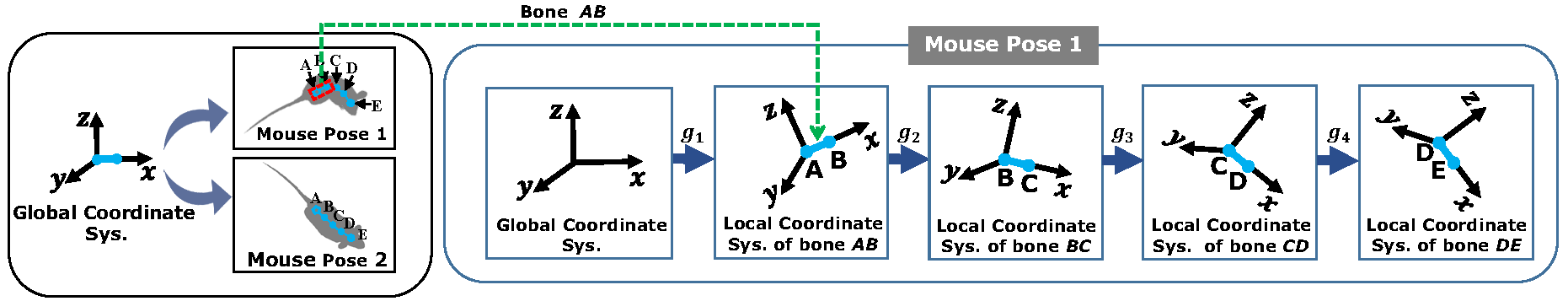}
\end{center}
\caption{The left figure shows two example poses of a mouse, where A, B, C, D and E are joints of the mouse spine. The global coordinate system is a pre-defined fixed coordinate system for reference purpose. The right figure demonstrates how to represent mouse pose 1, where the rigid transformation between two successive bones are described by rotation and translation. For example, the rigid transformation between bones $\protect\overrightarrow{\small{AB}}$ and $\protect\overrightarrow{BC}$ is described by $g_2 \in SE(3)$.}
\label{fig:coordinate}
\end{figure*}
As depicted in Figure~\ref{fig_skeleton}, a human is represented as a kinematic tree consisting of five kinematic chains: the spine and the four limbs, with a total of 42 DoFs. The numbers in the figure demonstrate the DoFs of the corresponding bones. A fish and a mouse are both represented as a single kinematic chain along the main spine with 25 and 12 DoFs, respectively.

Formally, the relative 3D geometry between two successive bones can be described by a 3D rigid transformation $g$ (mathematically, $g$ is an element of the Special Euclidean group $SE(3)$), which can be decomposed into a rotation and a translation. Take the mouse kinematic model as an example, as illustrated in Figure~\ref{fig:coordinate} we predefine a global coordinate system (see the left figure), and $g_1$ is defined as the rigid transformation of the first bone relative to the global reference system. For a mouse pose (e.g., \emph{mouse pose} 1 in the left figure of Figure~\ref{fig:coordinate}), a local coordinate system is attached to each of the bones (shown in the right figure of Figure~\ref{fig:coordinate}) such that the $x$-axis is aligned with the bone and the origin coincides with the start joint of the bone. A 3D rigid transformation $g$ relates the local coordinate systems of two successive bones. $g$ can be represented as a $4\times4$ matrix $g=\left(\begin{matrix}\mathrm{R} & \mathrm{\mathbf{t}} \\ 0 & 1\end{matrix}\right)$, with $\mathrm{R}$ being a $3\times3$ rotation matrix, and $\mathrm{\mathbf{t}}$ a 3D translation vector. More specifically, a joint with coordinates $\mathrm{\mathbf{x}} = (x,y,z)^\intercal$ w.r.t. coordinate system $i$ will have coordinates $\mathrm{\mathbf{x}}' = (x',y',z')^\intercal$ w.r.t coordinate system $i-1$ with
\begin{equation}
\left(\begin{matrix}
\mathrm{\mathbf{x}}' \\ 1
\end{matrix} \right)
=
\left(\begin{matrix}
\mathrm{R}_i & \mathrm{\mathbf{t}}_i \\
0 & 1
\end{matrix} \right)
\left(\begin{matrix}
\mathrm{\mathbf{x}} \\ 1
\end{matrix} \right)= g_i\left(\begin{matrix}
\mathrm{\mathbf{x}} \\ 1
\end{matrix} \right)
\end{equation}
Descending along a kinematic chain, the 3D coordinates of joint $m$ with respect to the global reference frame can be conveniently computed by forward kinematics, i,e, a sequence of successive 3D rigid transformations $g_1g_2 \cdots g_m$ acting on the origin. As such, the 3D coordinates of all joints along a kinematic chain can be fully characterized by a sequence of successive 3D rigid transformations.

\textbf{Bone Length Invariance} \quad
Taking into account the fact that bones are rigid objects with invariant lengths, there are no translational DoFs relating bones. As such, the kinematic chain framework naturally encodes the bone length invariance constraints and it suffices to consider the orientation of each bone along a chain.

\textbf{Rotational DoFs} \quad
After removing the 3 translational DoFs by considering the bone length invariance, we further empirically obtain the actual rotational DoFs of a skeletal joint. We first parameterize all the rotations with \emph{intrinsic} Euler angles (yaw, pitch, row). If a joint has a zero Euler angle for a principal axis across the entire training dataset, we take it to imply that there is no rotational DoF with respect to this axis.

The above procedures allow us to explicitly extract the actual DoF of each bone (as shown in Figure~\ref{fig_skeleton}) following the anatomical constraints. Thus, the dimensionality of the pose manifold is minimized, significantly reducing the search space for making predictions and improving the naturalness of the predicted pose. Next we present the different parameterization formalisms for the rotations within the kinematic chain representation framework.

\subsubsection{Axis angle parameterization of rotation}
Many existing works have adopted the axis angle parameterization for rotation~\cite{srnn,residualgru,AGED}. Given a rotation matrix $\mathrm{R}$, the axis parameters $\bm{\omega}$ are given by:
\begin{equation}\label{eqn:axisangles}
\begin{aligned}
\theta &= \arccos\left(\frac{\text{Tr}(\mathrm{R})-1}{2}\right),\\
\bm{\omega}& =
\frac{\theta}{2\sin\theta}\left(\begin{matrix}
\mathrm{R}(3,2)-\mathrm{R}(2,3)\\
\mathrm{R}(1,3)-\mathrm{R}(3,1)\\
\mathrm{R}(2,1)-\mathrm{R}(1,2)
\end{matrix}\right).
\end{aligned}
\end{equation}

Equation~(\ref{eqn:axisangles}) parameterizes rotation $\mathrm{R}$ as a rotation of angle $\theta$ about the unit axis $\hat{\bm{\omega}} = \frac{\bm{\omega}}{\norm{\bm{\omega}}}$. Formally, the axis angle parameters are elements of the Lie algebra $\mathfrak{so}(3)$ associated with the 3D rotation group $SO(3)$, and $\mathrm{R} \in SO(3)$.

The inverse transformation from $\bm{\omega}=\theta\hat{\bm{\omega}}=\theta\left(\begin{matrix}
\omega_1\\
\omega_2\\
\omega_3
\end{matrix}
\right)$ to $\mathrm{R}$ is given by the exponential map:
\begin{equation}\label{eqn:exponentialmap}
\begin{aligned}
\omega_\times &\myeq \theta\left(\begin{matrix}
0 & -\omega_3 & \omega_2 \\
\omega_3 & 0 & -\omega_1 \\
-\omega_2 & \omega_1 & 0
\end{matrix}
\right)\\
\exp:\mathfrak{so}(3)& \to SO(3)\\
\theta\omega_\times & \mapsto \mathrm{R}=\mathbb{1}+\sin\theta\omega_\times+(1-\cos\theta)\omega_\times^2.
\end{aligned}
\end{equation}
In existing literature such as~\cite{srnn,residualgru,erd}, parameterization using Equation~(\ref{eqn:axisangles}) is referred to as \emph{exponential map} representation.
\begin{figure*}[ht!]
\begin{center}
\includegraphics[width=1\linewidth]{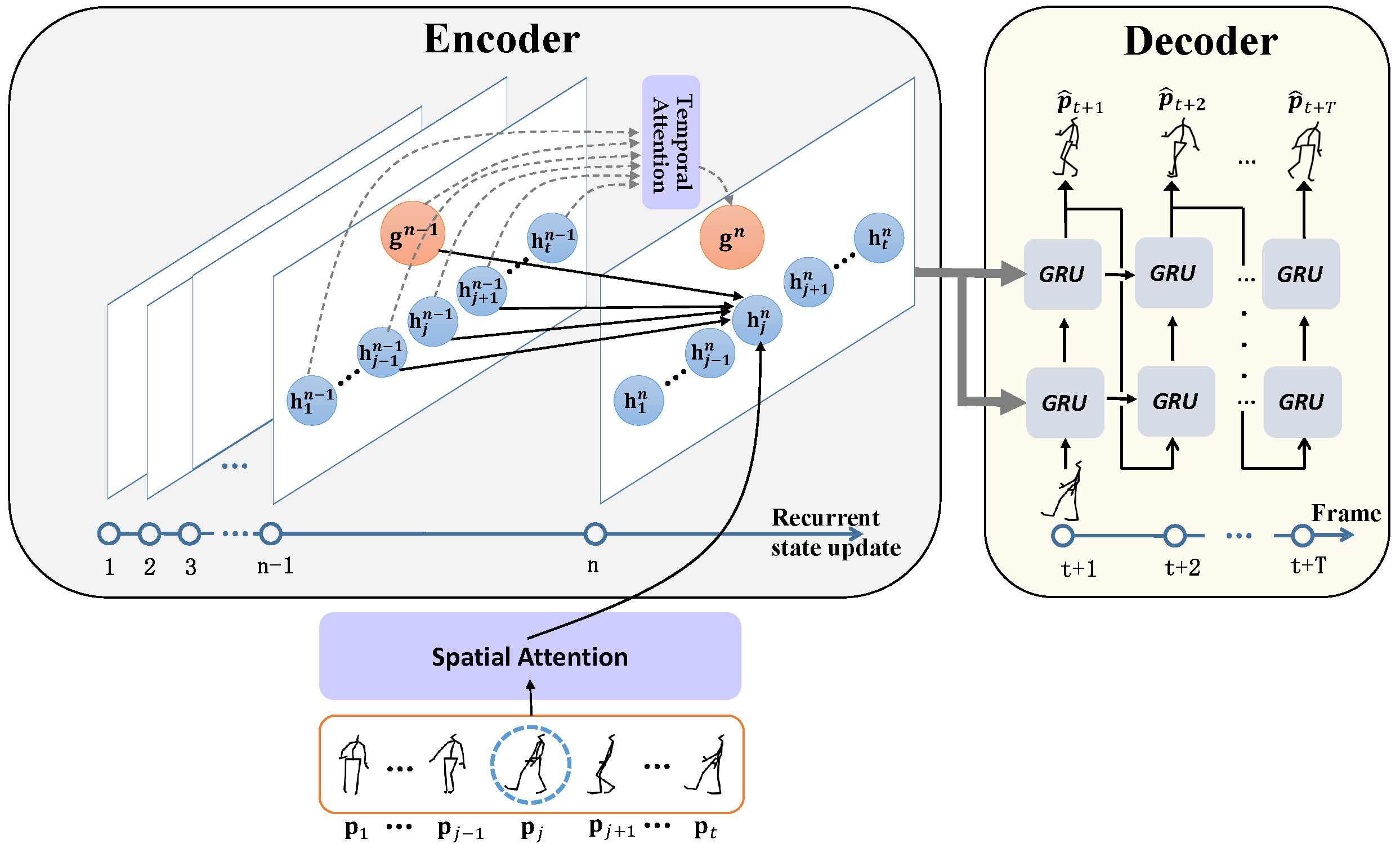}
\end{center}
\caption{The proposed neural network unfolded over recurrent steps. At a given recurrent step $n$, local hidden state ${\mathbf{h}}_j^n$ is
updated based on ${\mathbf{h}}_{j-1}^{n-1},~ {\mathbf{h}}_j^{n-1},~ {\mathbf{h}}_{j+1}^{n-1},~ {\mathbf{g}}^{n-1}$, and $\mathbf{p}_j$.
Global state $\mathbf{g}^{n}$ is updated based on
$\mathbf{g}^{n-1}$ and $\{\mathbf{h}_j^{n-1}\}_{i=1}^{t}$.}
\label{fig:network}
\vspace{-1em}
\end{figure*}
\subsubsection{Quaternion parameterization of rotation} 
An alternate rotation parameterization employing quaternions has been proposed in~\cite{pavllo2019modeling}. Quaternions are a number system developed by Sir William Rowan Hamilton. A quaternion can be represented as a 4-dimensional object $\mathbf{q}\in\mathbb{R}^4$. Each unit quaternion corresponds to a 3D rotation. Given a rotation of angle $\theta$ about the axis $\hat{\bm{\omega}}$, the quaternion representation can be derived as:
\begin{equation}\label{eqn:axisangletoquaternion}
\begin{aligned}
\mathbf{q} &= \left(\begin{matrix}
\cos \frac{\theta}{2}\\
\sin \frac{\theta}{2} \hat{\bm{\omega}}
\end{matrix}\right).
\end{aligned}
\end{equation}
It can be observed that $\mathbf{q}$ in the above corresponds to a 4-dimensional unit vector.

Converting a unit quaternion $\mathbf{q} = (q_1,q_2,q_3,q_4)^T$ to the axis angle representation is given by:
\begin{equation}\label{eqn:quaterniontoaxisangle}
\begin{aligned}
\theta &= 2\textrm{atan2} \left(\sqrt{q_2^2+q_3^2+q_4^2},q_1\right)\\
\hat{\bm{\omega}}&=\frac{(q_2,q_3,q_4)^T}{\sqrt{q_2^2+q_3^2+q_4^2}}.
\end{aligned}
\end{equation}
Mathematically, the unit quaternions forms a double covering of the the 3D rotation group $SO(3)$~\cite{hall2003lie}. This translates to the fact that an unit quaternion $\mathbf{q}$ and $-\mathbf{q}$ actually represents the same rotation, which can be a potential source of ambiguity. Quaternion representation also incurs the additional step of normalization, which introduces an additional source of difficulty for learning.

\subsubsection{Stiefel manifold parameterization of rotation} \label{sec:stiefel}
The $V_2(\mathbb{R}^3)$ Stiefel manifold, \emph{i.e.}, an ordered set of 2-tuple orthonormal 3D vectors offers another way to parameterize rotations. For a rotation $\mathrm{R} = \left(\begin{matrix}\mathbf{R}_1 & \mathbf{R}_2 &\mathbf{R}_3 \end{matrix}\right)$, the $V_2(\mathbb{R}^3)$ Stiefel manifold representation $\tilde{\mathrm{R}}$ is simply given by
\begin{equation}\label{eqn:stiefelmanifold}
\begin{aligned}
\tilde{\mathrm{R}} &= \left(\begin{matrix} \vertbar & \vertbar \\ \mathbf{R}_1 & \mathbf{R}_2 \\ \vertbar & \vertbar \end{matrix}\right)
\end{aligned}
\end{equation}
$\tilde{\mathrm{R}}$ uses 6 parameters to represent a rotation. Recovering the rotation $\mathrm{R}$ from the Stiefel manifold parameterization $\tilde{\mathrm{R}}$ can be simply performed with a cross product $\mathbf{R}_3 = \mathbf{R}_1 \times \mathbf{R}_2$.

\cite{zhou2019continuity} presented theoretical motivations for employing Stiefel manifold to represent rotations. The Stiefel manifold parameterization offers a globally continuous parameterization of 3D rotations, which offers empirical advantage for backpropagation training. It had been proven in~\cite{astey1987cobordism} that the real projective space in 3 dimensions $\mathbb{RP}^3$ (which is homeomorphic to the 3D rotation group $SO(3)$ does not admit an embedding in $\mathbb{R}^n$ for $n<5$. Topologically, this means that any parameterization of 3D rotations in a dimension smaller than 5 (which is the case for both axis angles and quaternions) would not be globally continuous. Inspired by~\cite{zhou2019continuity}, we apply the Stiefel manifold parameterization to our motion prediction task, motivated by the fact that this parameterization is globally continuous.

In addition, as we will discuss in subsection~\ref{sec_loss}, we define loss functions directly involving computations of the rotation matrices. The ease of recovering the rotation matrix from the Stiefel manifold again offers a key advantage over the axis angle or quaternion parameterizations which require more computationally expensive exponential operation. As such, both the global continuity and the reduced computational complexity in optimizing for the losses are desirable for model learning via backpropagation.

\section{Method}
\subsection{Method Overview}
\textbf{Problem Definition} \quad
Presented with an observed sequence of 3D skeletal poses $\langle \mathbf{p}_1, {\mathbf{p}}_2, \cdots , {\mathbf{p}}_t \rangle$, we are interested in predicting the future pose sequence $\langle {\mathbf{p}}_{t+1}, {\mathbf{p}}_{t+2}, \cdots \mathbf{p}_{t+T} \rangle$. Unlike \emph{action recognition} that predicts a one-dimensional action label, the output of motion prediction is a {high dimensional} pose sequence. The challenges also entail the complex nature of motion, as well as the subtle skeletal constraints to be considered for natural motion prediction.

Most existing methods simply employ conventional LSTM or GRU units to model motion contexts, which are known to have problems in capturing long-term dependencies~\cite{LSTM_problem}. Consequently, their predictions tend to converge to motionless states in the long-term (shown in Figure~\ref{fig:deficiency}, Figure~\ref{fig:walking}, and the supplementary video). 
To tackle this issue, we propose a novel \emph{AHMR} network to effectively model global and local motion contexts. 

Another issue with existing approaches is that their representation of pose inputs may be suboptimal. Representing ${\mathbf{p}}_j$ as raw 3D joint coordinates~\cite{predictionandclassification} neglects the dependencies between joints whereas the axis angle~\cite{representation2,srnn,Residual} or quaternion~\cite{pavllo2019modeling} are subject to discontinuity in the underlying representation space. Consequently, this leads to inaccurate predictions or severe body distortions (shown in Figure~\ref{fig:deficiency}, Figure~\ref{fig:walking}, and Subsection~\ref{sec:quantitative}). To this end, we perform extensive experiments over the different pose parameterizations discussed in Section~\ref{sec:pose} to study their feasibility and performance in the motion prediction task. Furthermore, we also examine various loss functions including a geodesic loss and a forward kinematics loss, a smooth L1 loss as well as the commonly adopted L2 loss. The comprehensive investigation over the different combinations of pose representations and loss functions is reported in subsection~\ref{sec:poserep&loss}.

In the following, we present the details of the \emph{AHMR} network and two geometrically motivated loss functions that we explore for the motion prediction task.

\begin{figure*}[t]
\begin{center}
\includegraphics[width=0.9\linewidth]{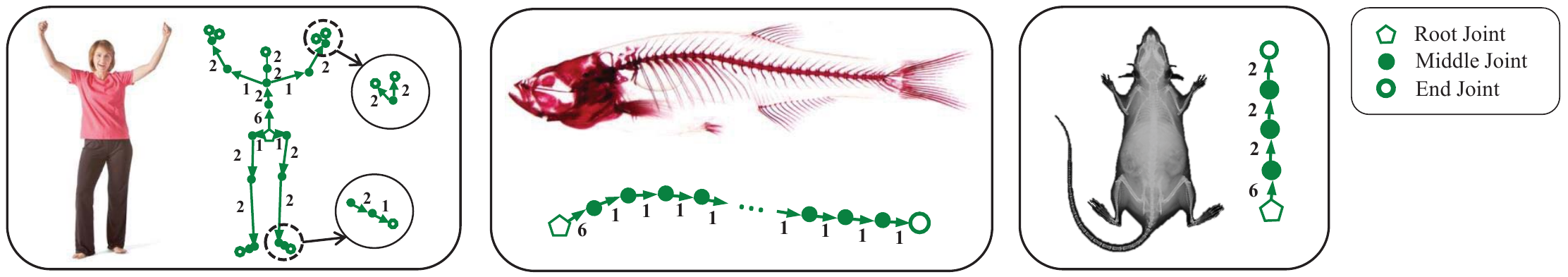}
\end{center}
\caption{The kinematic chain structures of human, fish, and mouse, respectively. The arrows correspond to bones, while the numbers near the bones demonstrate their DoFs. The first bones of all the three skeletons are of 6 DoFs, while all other bones in the fish and mouse skeleton have 1 DoF and 2 DoFs, respectively. The first bone amounts to the bone located in the main spine and starts from the root joint.}
\label{fig_skeleton}
\end{figure*}

\begin{figure*}[ht!]
\begin{center}
\begin{subfigure}{.5\linewidth}
	\includegraphics[height=.2\textheight]{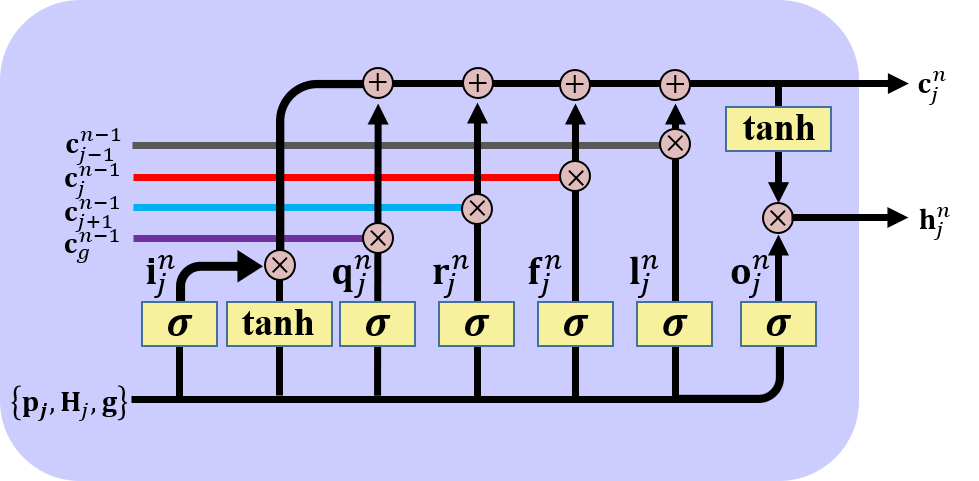}
	\caption{}
	\label{fig:h_update}
\end{subfigure}
\hspace{4em}
\begin{subfigure}{.4\linewidth}
	\includegraphics[height=.2\textheight]{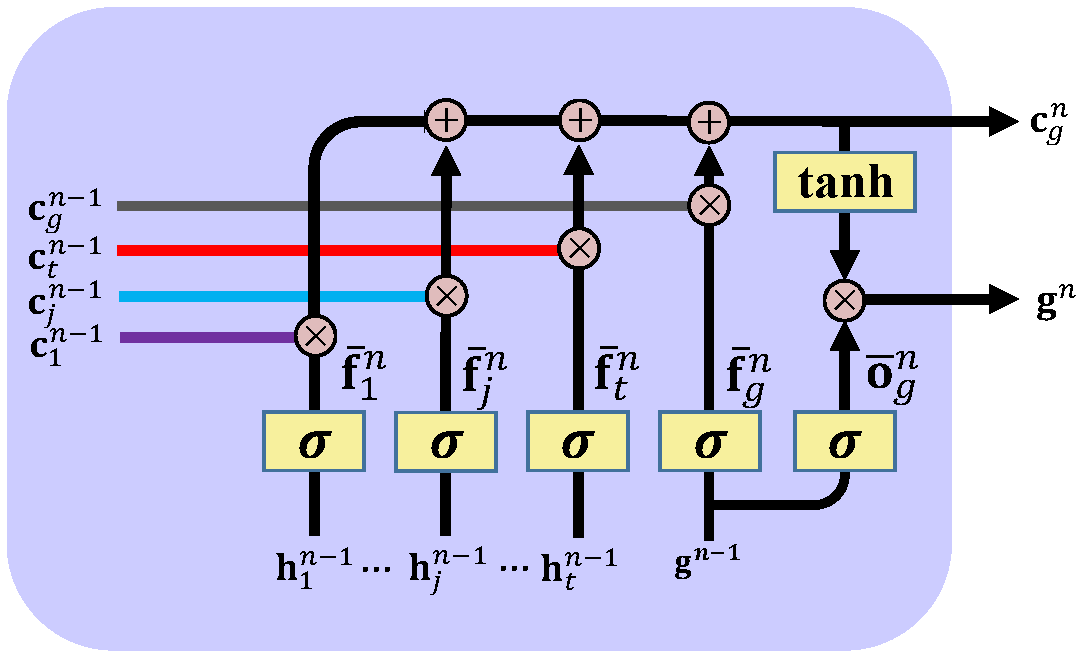}
	\caption{}
	\label{fig:g_update}
\end{subfigure}%
\caption{A display of the neural gates. The left figure visualizes the gates for updating local state $\mathbf{h}_j^{n}$ and its corresponding cell state ${\mathbf{c}}_j^n$, while the right figure visualizes the gates for updating global state $\mathbf{g}^n$ and its corresponding cell state $\mathbf{c}_g^n$.}
\label{fig_gatevisualization}
\end{center}
\end{figure*}

\subsection{AHMR Network for Encoding Motion Context}
\label{AHMR} 
In this subsection, we introduce how to model motion contexts and generate future pose sequence using our proposed network structure.

In conventional RNN-based motion prediction methods~\cite{erd,srnn,residualgru}, the observed poses $\langle {\mathbf{p}}_1, \cdots, \mathbf{p}_t\rangle$ are successively input into an encoder network and the hidden state $\mathbf{h}_{j}$ is updated upon reading each pose $\mathbf{p}_j$, which is given by:
\begin{equation}\label{eq:conventionalrnn}
\mathbf{h}_{j} = \textrm{RNN}( \mathbf{p}_j, \mathbf{h}_{j-1})
\end{equation}
The last hidden state $\mathbf{h}_{t}$ is taken as the final motion context to predict future poses. It has been pointed out that $\mathbf{h}_{t}$ has difficulties in modeling \emph{long-term} dynamics~\cite{LSTM_problem}. Recently,~\cite{mhu} proposes to address this by allowing $\mathbf{h}_{t}$ to have direct access to all historical poses $\{\mathbf{p}_j\}_{j=1}^n$. Unfortunately, this approach heavily relies on the individual poses and fails to consider the motion contexts hidden in consecutive poses.

Inspired by this, we propose a new architecture termed \emph{AHMR}. As shown in Figure~\ref{fig:network}, the proposed model consists of an \emph{encoder} and a \emph{decoder}. The encoder learns a two-level motion context for the entire input pose sequence, which is subsequently passed to the decoder for translating the context and outputting the future pose sequence.

\textbf{Encoder} \quad In our encoder, motion contexts $\mathbf{S}$ are jointly modeled by local states $\{\mathbf{h}_j\}_{j=1}^t$ for individual frames and a global state $\mathbf{g}$ for the entire sequence. Formally, $\mathbf{S}=<\mathbf{h}_1,\mathbf{h}_2,\cdots, \mathbf{h}_t, \mathbf{g}>$. At each recurrent step $n$, the $j^{th}$ frame updates its local state $\mathbf{h}_j^n$ by exchanging information with its neighboring local states $\mathbf{h}_{j+1}^{n-1}$ and $\mathbf{h}_{j-1}^{n-1}$, and additionally with the global state $\mathbf{g}^n$. As the number of recurrent steps increases, the motion context $\mathbf{h}_j^n$ of each frame is incrementally enriched by accessing the local states of adjacent frames and the global state. 
The optimum number of recurrent steps is determined by the task, while in conventional sequential RNN models the number of recurrent steps is determined by the number of frames in the input sequence.

Typically, different parts of the body engage differently in a motion. For example, actions such as walking would primarily involve synchronized motion of the limbs whereas smoking would mainly involve movements of a single arm. Motivated by this, we propose to integrate an spatial attention module into the \emph{AHMR} encoder for modeling such spatial dependencies. The velocity and acceleration of a joint $i$ are important indicators in predicting its future motion and thus we use these heuristics in assigning weights $\beta^i$ (velocity) and $\gamma^i$ (acceleration) to joint $i$.
\begin{equation} \label{eqn:spatialattention}
\begin{aligned}
\bar{\mathbf{p}}_j^i&=\beta_j^i\mathbf{p}_j^i+\gamma_j^i\mathbf{p}_j^i\\
\beta_j^i &= \frac{\exp \norm{\mathbf{p}_j^i-\mathbf{p}_{j-1}^i}}{\sum_i \exp \norm{\mathbf{p}_j^i-\mathbf{p}_{j-1}^i}}\\
\gamma_j^i &= \frac{\exp \norm{\mathbf{p}_j^i-2\mathbf{p}_{j-1}^i+\mathbf{p}_{j-2}^i}}{\sum_i \exp \norm{\mathbf{p}_j^i-2\mathbf{p}_{j-1}^i+\mathbf{p}_{j-2}^i}}
\end{aligned}
\end{equation}
where $\mathbf{p}_j^i$ denotes the location of the $i^{th}$ joint in the $j^{th}$ frame. $\beta^i$ models velocity while $\gamma^i$ measures acceleration.

The network is initialized with:
\begin{equation}
\begin{aligned}
\mathbf{h}_j^0&=W_{\mathrm{in}}{\bar{\mathbf{p}}}_{j}+\mathbf{b}_{\mathrm{in}}\\
\mathbf{g}^0&=\frac{1}{t}\sum\limits_{j=1}^{t}{\mathbf{h}}_j^{0}\\
\mathbf{c}_j^0&=\mathbf{c}_g^0=\mathbb{1},
\end{aligned}
\end{equation}
where ${\mathbf{c}}_j^0,{\mathbf{c}}_h^0$ denotes the initial \emph{cell state} associated with $\mathbf{h}_j^0$ and $\mathbf{g}^0$ respectively. Matrix $W_{\mathrm{in}}$ and biases ${\mathbf{b}}_{\mathrm{in}}$ are parameters to be learned.

\textbf{Update frame-level state $\mathbf{h}_j^n$} \quad
Subsequently, for a given \emph{recurrent step} $n$, the local motion context $\mathbf{h}_j^{n}$ is calculated based on $\mathbf{h}_{j-1}^{n-1}, \mathbf{h}_{j}^{n-1}, \mathbf{h}_{j+1}^{n-1}, \mathbf{g}^{n-1},$ and $\mathbf{p}_j$. Specifically, the state transition process from $\mathbf{h}_j^{n-1}$ to $\mathbf{h}_j^{n}$ is formulated as below. For clarity, we also visualize the process in Fig.~\ref{fig:h_update}.
\begin{equation}\label{eq_updateh}
\begin{aligned}
\mathbf{H}_j^{n-1}&=(\mathbf{h}_{j-1}^{n-1};\mathbf{h}_{j}^{n-1};\mathbf{h}_{j+1}^{n-1}) \\
\mathbf{f}_j^n &= \sigma(U_f\,\mathbf{p}_j+W_f\,\mathbf{H}_j^{n-1}+Z_f\,\textbf{g}^{n-1}+\mathbf{b}_f) \\
\mathbf{l}_j^n &= \sigma(U_l\,\mathbf{p}_j+W_l\,\mathbf{H}_j^{n-1}+Z_l\,\mathbf{g}^{n-1}+\mathbf{b}_l)\\
\mathbf{r}_j^n &= \sigma(U_r\,\mathbf{p}_j+W_r\,\mathbf{H}_j^{n-1}+Z_r\,\mathbf{g}^{n-1}+\mathbf{b}_r)\\
\mathbf{q}_j^n &= \sigma(U_q\,\mathbf{p}_j+W_q\,\mathbf{H}_j^{n-1}+Z_q\,\mathbf{g}^{n-1}+\mathbf{b}_q)\\
\mathbf{i}_j^n &= \sigma(U_i\,\mathbf{p}_j+W_i\,\mathbf{H}_j^{n-1}+Z_i\,\mathbf{g}^{n-1}+\mathbf{b}_i)\\
\mathbf{o}_j^n &= \sigma(U_o\,\mathbf{p}_j+W_o\,\mathbf{H}_j^{n-1}+Z_o\,\mathbf{g}^{n-1}+\mathbf{b}_o)\\
\tilde{\textbf{c}}_j^n &= \tanh(U_c \, \mathbf{p}_j+W_c\,\mathbf{H}_j^{n-1}+Z_c\,\mathbf{g}^{n-1}+\mathbf{b}_c)\\
\textbf{c}_j^n &= \mathbf{l}_j^n\odot \textbf{c}_{j-1}^{n-1} + \mathbf{f}_j^n \odot \textbf{c}_j^{n-1} + \mathbf{r}_j^n \odot \textbf{c}_{j+1}^{n-1} \\
& \quad \quad \quad + \mathbf{q}_j^n \odot \textbf{c}_g^{n-1} + \mathbf{i}_j^n\odot \tilde{\textbf{c}}_j^n\\
\mathbf{h}_j^{n} &= \mathbf{o}_j^n \odot \tanh(\textbf{c}_j^n)
\end{aligned}
\end{equation}

As illustrated in Equation~(\ref{eq_updateh}), $\mathbf{H}_j^{n-1}$ is the concatenation of a neighboring context window. A total of 4 \emph{forget} gates, namely $\mathbf{f}_j^n$ (forget gate), $\mathbf{l}_j^n$ (left forget gate), $\mathbf{r}_j^n$ (right forget gate), $\mathbf{q}_j^n$ (global forget gate), are computed, which respectively control the information flows from the current cell state ${\mathbf{c}}_{j}^{n-1}$, left cell state ${\mathbf{c}}_{j-1}^{n-1}$, right cell state $\mathrm{\mathbf{c}}_{j+1}^{n-1}$, and global cell state $\mathrm{\mathbf{c}}_g^{n-1}$ to cell state ${\mathbf{c}}_j^n$. $\mathbf{i}_j^n$ denotes the \emph{input gate} from the estimated cell state $\tilde{\mathbf{c}}_j^n$ to ${\mathbf{c}}_j^n$, while $\mathbf{o}_j^n$ is the \emph{output gate} from ${\mathbf{c}}_j^n$ to $\mathbf{h}_j^{n}$. Cell state ${\mathbf{c}}_j^n$ is obtained by a Hadamard product of the five gates and the corresponding cells. $\mathbf{h}_j^{n}$ is given by the product of the \emph{output} gate $o_j^n$ and the $\tanh$ activated cell state ${\mathbf{c}}_j^n$. Matrices $U_k, W_k, Z_k$ and biases ${\mathbf{b}}_k$ are parameters to be learned where $k \in \{f,l,r,q,i,o\}$.

\textbf{Update sequence-level state $\mathbf{g}^n$} \quad
In updating the global motion context, we incorporate a temporal attention mechanism over the local motion contexts as given in Equation~(\ref{eqn:temporalattention}). This sets the weights $\alpha_j$ of motion context $\mathbf{h}_j^{n-1}$ according to its similarity with the motion context $\mathbf{h}_t^{n-1}$ of the last observed frame.
\begin{equation} \label{eqn:temporalattention}
\begin{aligned}
\bar{\mathbf{g}}&=\sum\limits_{j=1}^{t}\alpha_j\mathbf{h}_j^{n-1}\\
\alpha_j &= \frac{\exp(\mathbf{h}_j^{n-1} \cdot \mathbf{h}_t^{n-1})}{\sum_j \exp(\mathbf{h}_j^{n-1} \cdot \mathbf{h}_t^{n-1})}
\end{aligned}
\end{equation}

At a given \emph{recurrent step} $n$, the state transition process from $\mathbf{g}^{n-1}$ to $\mathbf{g}^n$ is given by Equation~(\ref{eq_updateG}). Similarly, we visualize this process in Fig.~\ref{fig:g_update}.
\begin{equation}\label{eq_updateG}
\begin{aligned}
\bar{\mathbf{g}}&=\sum\limits_{j=1}^{t}\alpha_j{\mathbf{h}_j^{n-1}}\\
\bar{\mathbf{f}}_j^n &= \sigma(\bar{W}_f\,\mathbf{h}_j^{n-1}+\bar{Z}_f\,\mathbf{g}^{n-1}+\bar{\mathbf{b}}_f) \\
\bar{\mathbf{f}}_g^n &= \sigma(\bar{W}_g\,\bar{\mathbf{g}}+\bar{Z}_g\,\mathbf{g}^{n-1}+\bar{\mathbf{b}}_g) \\
\bar{\mathbf{o}}_g^n &= \sigma(\bar{W}_o\,\bar{\mathbf{g}}+\bar{Z}_o\,\mathbf{g}^{n-1}+\bar{\mathbf{b}}_o)\\
\mathbf{c}_g^n &= \sum_{j=1}^{t} \bar{\mathbf{f}}_j^{n} \odot \mathbf{c}_j^{n-1} + \bar{\mathbf{f}}_g^n \odot \mathbf{c}_g^{n-1} \\
\mathbf{g}^{n} &= \bar{\mathbf{o}}_g^n \odot \tanh(\mathbf{c}_g^n)
\end{aligned}
\end{equation}
As shown in Equation~(\ref{eq_updateG}), $\bar{\mathbf{f}}_j^n$ and $\bar{\mathbf{f}}_g^n$ are the respective \emph{forget} gates that filter information from cell states ${\mathbf{c}}_j^{n-1}$ and ${\mathbf{c}}_g^{n-1}$ to cell state ${\mathbf{c}}_g^n$.
Output gate $\bar{\mathbf{o}}_g^n$ controls information flow from cell state $\mathbf{c}_g^n$ to the final global state $\mathbf{g}^{n}$. $\mathbf{c}_g^n$ is obtained by the Hadamard product of the two gates and the corresponding cells, while $\mathbf{g}^{n}$ is given by the Hadamard product of the \emph{output} gate $\bar{o}_i^n$ and the $\tanh$ activated cell state $\mathbf{c}_g^n$. Matrices $\bar{W}_k,\bar{Z}_k$ and biases $\bar{\mathbf{b}}_k$ with index $k \in \{f, g, o\}$ are parameters to be learned.

\textbf{Decoder} \quad Our encoder learns a two-level (frame-level and sequence-level) motion context for the entire input sequence, which is subsequently passed into the decoder to translate the context and output future motion sequence. As displayed in Figure~\ref{fig:network}, our decoder engages a two-layer stacked GRU network. The hidden state of the first layer is set as $\mathbf{g}^n$, while the hidden state of the second layer is configured as $\mathbf{h}_t^n + \mathbf{g}^n$. The decoder is executed following the directed links shown in Figure~\ref{fig:network}, where the predicted pose for frame $t+i$ is fed back as input to recursively generate pose prediction for frame $t+i+1$.

\textbf{Discussions} \quad The default number of neighboring context window size in \emph{AHMR} is set to 3, namely for recurrent step $n$, ${\mathbf{s}}_i^n$ only exchanges information with its immediate neighbors ${\mathbf{s}}_{i-1}^n$ and ${\mathbf{s}}_{i+1}^n$. Although the connections between the global state and all local states may speed up information exchange between nonconsecutive local states, it is interesting to see the effect of enlarging or reducing neighboring window size. We empirically investigate the influence of different window sizes and effect of different numbers of recurrent steps in subsection~\ref{sec:ahmrsettings}.

\subsection{Loss Function} \label{sec_loss}
A common evaluation metric for motion prediction accuracy is the mean angle error. Geometrically, this metric concerns the geodesic distance on the SO(3) manifold, which measures the smallest angular difference between two rotations. Most existing works such as~\cite{erd,srnn,residualgru} adopted a L2 loss function for the axis angle parameters, which unfortunately is not a good metric for measuring the difference between rotations. In part this is due to the discontinuity in the parameterization. For instance, consider three rotations given in axis angle parameters as $\bm{\omega}_1 = \left(0.01,0,0\right)^T$, $\bm{\omega}_2 = \left(0.5,0,0\right)^T$, $\bm{\omega}_3 = \left(6.27,0,0\right)^T$. We have $\norm{\bm{\omega}_1-\bm{\omega}_2}_2<\norm{\bm{\omega}_1-\bm{\omega}_3}_2$ whereas in actuality the associated rotations $\mathrm{R}_1$ and $\mathrm{R}_3$ are much more similar.

On a more fundamental level, a rotation parameterization is a mapping from $(\text{SO(3)}, \text{geodesic})$ to $(\mathcal{X}, d_{\mathcal{X}})$ where $d_{\mathcal{X}}$ is the metric over the parameter space $\mathcal{X}$. Taking $d_{\mathcal{X}}$ as the L2 or Euclidean distance $\lVert . \rVert_2$ inevitably introduces a metric distortion of the geodesic distance for all considered parameterization schemes, which are detrimental to the prediction accuracy.\footnote{E.g. consider rotations $\mathrm{R}_1 = \left(\begin{matrix} 0.80 & -0.28 & 0.53 \\ 0.46 & 0.85 & -0.25 \\ -0.38 & 0.44 & 0.81 \end{matrix}\right)$, $\mathrm{R}_2 = \left(\begin{matrix} -0.71 & 0.52 & 0.48 \\ 0.71 & 0.46 & 0.54 \\ 0.05 & 0.72 & -0.69 \end{matrix}\right)$, and $\mathrm{R}_3 = \left(\begin{matrix} -0.61 & 0.36 & 0.70 \\ 0.50 & -0.51 & 0.70 \\ 0.61 & 0.78 & 0.14 \end{matrix}\right)$. For the axis angle, $\text{geodesic}(\mathrm{R}_1,\mathrm{R}_3)/\norm{\bm{\omega}_1-\bm{\omega}_3}_2= 0.997,$ and $\text{geodesic}(\mathrm{R}_2,\mathrm{R}_3)/\norm{\bm{\omega}_2-\bm{\omega}_3}_2= 0.684.$
\\For the Stiefel manifold, $\text{geodesic}(\mathrm{R}_1,\mathrm{R}_3)/\norm{\tilde{\mathrm{R}}_1-\tilde{\mathrm{R}}_3}_2= 1.006,$ and  $\text{geodesic}(\mathrm{R}_2,\mathrm{R}_3)/\norm{\tilde{\mathrm{R}}_2-\tilde{\mathrm{R}}_3}_2= 0.929.$ 
\\In either case, such distortions result in misalignment of the training loss and the mean angle error metric.}

\textbf{Geodesic Loss}\quad In account of this, we utilize the {geodesic loss} function. For a given set of rotations $\{\mathrm{R}\}_{i=1}^N$ and $\{\hat{\mathrm{R}}\}_{i=1}^N$, the geodesic loss is defined as the squared geodesic distance:
\begin{equation}\label{eqn:geodesic}
\begin{aligned}
\mathcal{L}_{\text{geodesic}}(\mathrm{R}, \hat{\mathrm{R}})&=\left[\arccos 
\left(\frac{\text{Tr}(\mathrm{R}\hat{\mathrm{R}}^T)-1}{2}\right)\right]^2.
\end{aligned}
\end{equation}
This geodesic loss is parameterization independent and characterizes the squared shortest distance separating two rotations on the SO(3) manifold. 

\textbf{Forward Kinematics Loss}\quad On top of the geodesic loss, we further examine another geometrically meaningful and parameterization independent loss, namely the {forward kinematics loss} which measures the mean position error of the 3D joint coordinates characterized by the kinematics chain representation. Along a kinematic chain, the forward kinematics procedure recovers the coordinates of the $k^{th}$ joint:
\begin{equation}\label{eqn:forwardkinematics}
\begin{aligned}
\Upsilon_k=\prod_{i=0}^{k-1} \mathrm{R}_i\bm{T}_k+\Upsilon_{k-1}
\end{aligned}
\end{equation}
where $\bm{T}_k$ is the translation vector characterizing the reference position of the $k^{th}$ bone. The forward kinematics loss is computed as:
\begin{equation}\label{eqn:forwardkinematicsloss}
\begin{aligned}
\mathcal{L}_{\text{kinematics}}\left(\{\mathrm{R}_i\}_{i=0}^{m-1}, \{\hat{\mathrm{R}}_i\}_{i=0}^{m-1}\right)=\sum_{i=1}^{m}\norm{\Upsilon_i -\hat{\Upsilon}_i}^2
\end{aligned}
\end{equation}
The kinematic chain follows a hierarchical structure, \emph{i.e.} orientations of different bones have a varying significance when computing the joint positions. Technically, the position of the $i^{th}$ joint is computed based on the  position of the $i-1^{th}$ joint using rotations. The forward kinematics loss naturally takes into account the higher importance of rotations nearer to the chain root. It is thus more suitable when we are interested in benchmarking our performance via a mean position error metric. On the other hand, the geodesic loss or L2 loss assigns an equal weight to each bone orientation, and are thus more suitable for the mean angle error metric.

We would like to point out that the forward kinematic chain loss is distinct from the conventional L2 loss. The L2 loss directly computes the mean angle error between the ground truth and the prediction. In contrast, the forward kinematic chain loss unfolds all the rotations, considering that the bones possess a chain structure and a bone rotation affects its subsequent bones. To the best of our knowledge, the forward kinematics loss has not been explored in the motion prediction task yet.

In subsection~\ref{sec:poserep&loss}, we examine the empirical effectiveness of different loss functions, including the geodesic loss, forward kinematics loss, and the commonly adopted L2 loss and smooth L1 loss.

\begin{table*}[ht!]
\resizebox{1\textwidth}{!}
{
\begin{tabular}{|c|cccc|cccc|cccc|cccc|}
\hline
Time(ms) & 80 & 160 & 320 & 400 & 80 & 160 & 320 & 400 & 80 & 160 & 320 & 400 & 80 & 160 & 320 & 400 \\ \hline\hline
& \multicolumn{4}{c|}{Walking} & \multicolumn{4}{c|}{Eating} & \multicolumn{4}{c|}{Smoking} & \multicolumn{4}{c|}{Discussion} \\ \hline
ERD~\cite{erd} & 0.94 & 1.19 & 1.58 & 1.76 & 1.28 & 1.46 & 1.67 & 1.81 & 1.67 & 1.96 & 2.32 & 2.41 & 2.22 & 2.38 & 2.58 & 2.69 \\ \hline
SRNN~\cite{srnn} & 0.81 & 0.94 & 1.17 & 1.31 & 0.99 & 1.16 & 1.39 & 1.50 & 1.38 & 1.60 & 1.89 & 2.02 & 1.16 & 1.40 & 1.75 & 1.85 \\ \hline
Zero-Velocity & 0.39 & 0.68 & 0.99 & 1.15 & 0.27 & 0.48 & 0.73 & 0.86 & 0.32 & 0.60 & 1.00 & 1.11 & 0.31 & 0.67 & 0.97 &1.04 \\ \hline
ResGRU~\cite{residualgru} & 0.29 & 0.49 & 0.71 & 0.78 & 0.25 & 0.42 & 0.68 & 0.83 & 0.32 & 0.60 & 1.00 & 1.11 & 0.31 & 0.69 & 1.03 & 1.12 \\ \hline
QuaterNet~\cite{pavllo2019modeling} & 0.21 & 0.34 & 0.56 & 0.62 & 0.20 & 0.35 & 0.58 & 0.70 & 0.25 & 0.47 & 0.93 & 0.90 & 0.26 & 0.60 & 0.85 & 0.93 \\ \hline
CEM~\cite{li2018convolutional} & 0.33 & 0.54 & 0.68 & 0.73 & 0.22 & 0.36 & 0.58 & 0.71 & 0.26 & 0.49 & 0.96 & 0.92 & 0.32 & 0.67 & 0.94 & 1.01 \\ \hline
AGED~\cite{AGED} & 0.22 & 0.36 & 0.55 & 0.67 & 0.17 & 0.28 & 0.51 & 0.64 & 0.27 & 0.43 & 0.82 & 0.84 & 0.27 & 0.56 & 0.76 & 0.83 \\ \hline
Traj-GCN~\cite{mao2019learning} & 0.18 & 0.31 & 0.49 & 0.56 & \textbf{0.16} & 0.29 & 0.50 & 0.62 & 0.22 & 0.41 & 0.86 & 0.80 & \textbf{0.20} & 0.51 & 0.77 & 0.85 \\ \hline
DMGNN~\cite{li2020dynamic} & 0.18 & 0.31 & 0.49 & 0.58 & 0.17 & 0.30 & 0.49 & 0.59 & 0.22 & 0.39 & 0.81 & 0.77 & 0.26 & 0.65 & 0.92 & 0.99 \\ \hline
Transformer~\cite{aksan2020spatio} & 0.25 & 0.42 & 0.67 & 0.79 & 0.21 & 0.32 & 0.54 & 0.68 & 0.26 & 0.49 & 0.94 & 0.90 & 0.31 & 0.67 & 0.95 & 1.04 \\ \hline
AHMR (Ours) & \textbf{0.17} & \textbf{0.29} & \textbf{0.45} & \textbf{0.52} & 0.21 & \textbf{0.26} & \textbf{0.47} & \textbf{0.57} & \textbf{0.21} & \textbf{0.38} & \textbf{0.76} & \textbf{0.76} & \textbf{0.20} & \textbf{0.49} & \textbf{0.71} & \textbf{0.78} \\ \hline\hline

& \multicolumn{4}{c|}{Directions} & \multicolumn{4}{c|}{Greeting} & \multicolumn{4}{c|}{Phoning} & \multicolumn{4}{c|}{Posing} \\ \hline
Zero-Velocity & 0.25 & 0.44 & 0.61 & 0.68 & 0.80 & 1.23 & 1.81 & 1.87 & 0.80 & 1.23 & 1.81 & 1.87 & 0.32 & 0.63 & 1.16 & 1.45 \\ \hline
ResGRU~\cite{residualgru} & 0.41 & 0.64 & 0.80 & 0.92 & 0.57 & 0.83 & 1.45 & 1.60 & 0.59 & 1.06 & 1.45 & 1.60 & 0.45 & 0.85 & 1.34 & 1.56 \\ \hline
CEM~\cite{li2018convolutional} & 0.39 & 0.60 & 0.80 & 0.91 & 0.51 & 0.82 & 1.21 & 1.38 & 0.59 & 1.13 & 1.51 & 1.65 & 0.29 & 0.60 & 1.12 & 1.37 \\ \hline
Traj-GCN~\cite{mao2019learning} & {0.26} & {0.45} & 0.70 & 0.79 & {0.35} & {0.61} & {0.96} & {1.13} & {0.53} & {1.02} & 1.32 & 1.45 & 0.23 & 0.54 & 1.26 & 1.38 \\ \hline
DMGNN~\cite{li2020dynamic} & {\bf 0.25} & {\bf 0.44} & {\bf 0.65} & {\bf 0.71} & {0.36} & {0.61} & {0.94} & {\bf 1.12} & {0.52} & {0.97} & {1.29} & {1.43} & {0.20} & {0.46} & {1.06} & {1.34} \\ \hline
AHMR (Ours) & 0.28 & 0.55 & 0.66 & 0.74 & {\bf 0.33} & {\bf 0.59} & {\bf 0.90} & 1.18 & {\bf 0.39} & {\bf 0.93} & {\bf 1.25} & {\bf 1.37} & {\bf 0.19} & {\bf 0.44} & {\bf 1.10} & {\bf 1.29} \\ \hline\hline

& \multicolumn{4}{c|}{Purchases} & \multicolumn{4}{c|}{Sitting} & \multicolumn{4}{c|}{Sitting Down} & \multicolumn{4}{c|}{Taking Photo} \\ \hline 
Zero-Velocity & 0.72 & 1.03 & 1.46 & 1.49 & 0.43 & 1.12 & 1.41 & 1.58 & 0.27 & 0.54 & 0.93 & 1.05 & 0.22 & 0.47 & 0.78 & 0.89 \\ \hline
ResGRU~\cite{residualgru} & 0.58 & 0.79 & 1.08 & 1.15 & 0.41 & 0.68 & 1.12 & 1.33 & 0.47 & 0.88 & 1.37 & 1.54 & 0.28 & 0.57 & 0.90 & 1.02 \\ \hline
CEM~\cite{li2018convolutional} & 0.63 & 0.91 & 1.19 & 1.29 & 0.39 & 0.61 & 1.02 & 1.18 & 0.41 & 0.78 & 1.16 & 1.31 & 0.23 & 0.49 & 0.88 & 1.06 \\ \hline
Traj-GCN~\cite{mao2019learning} & 0.42 & 0.66 & 1.04 & 1.12 & 0.29 & 0.45 & 0.82 & 0.97 & {0.30} & {0.63} & {0.89} & {1.01} & {\bf 0.15} & {0.36} & {0.59} & {0.72} \\ \hline
DMGNN~\cite{li2020dynamic} & 0.41 & 0.61 & 1.05 & 1.14 & 0.26 & 0.42 & {\bf 0.76} & 0.97 & {0.32} & {0.65} & {0.93} & {1.05} & {\bf 0.15} & {\bf 0.34} & {0.58} & {0.71} \\
\hline
AHMR (Ours) & {\bf 0.38} & {\bf 0.61} & {\bf 1.01} & {\bf 1.09} & {\bf 0.22} & {\bf 0.41} & 0.77 & {\bf 0.89} & {\bf 0.30} & {\bf 0.61} & {\bf 0.85} & {\bf 0.99} & 0.16 & 0.35 & {\bf 0.55} & {\bf 0.69} \\ \hline\hline

& \multicolumn{4}{c|}{Waiting} & \multicolumn{4}{c|}{Walking Dog} & \multicolumn{4}{c|}{Walking Together} & \multicolumn{4}{c|}{Average} \\ \hline
Zero-Velocity & 0.27 & 0.49 & 0.96 & 1.12 & 0.60 & 0.96 & 1.27 & 1.33 & 0.33 & 0.60 & 0.96 & 1.03 & 0.42 & 0.74 & 1.12 & 1.20 \\ \hline
ResGRU~\cite{residualgru} & 0.32 & 0.63 & 1.07 & 1.26 & 0.52 & 0.89 & 1.25 & 1.40 & 0.27 & 0.53 & 0.74 & 0.79 & 0.40 & 0.69 & 1.04 & 1.18 \\ \hline
CEM~\cite{li2018convolutional} & 0.30 & 0.62 & 1.09 & 1.30 & 0.59 & 1.00 & 1.32 & 1.44 & 0.27 & 0.52 & 0.71 & 0.74 & 0.38 & 0.68 & 1.01 & 1.13 \\ \hline
Traj-GCN~\cite{mao2019learning} & 0.23 & 0.50 & 0.92 & 1.15 & 0.46 & 0.80 & {1.12} & {1.30} & {0.15} & 0.35 & 0.52 & {0.57} & {0.27} & 0.53 & 0.85 & 0.96 \\ \hline
DMGNN~\cite{li2020dynamic} & {0.22} & {\bf 0.49} & {0.88} & {1.10} & {0.42} & {\bf 0.72} & {1.16} & {1.34} & {0.15} & {0.33} & {0.50} & {0.57} & {0.27} & {0.52} & {0.83} & {0.95} \\ \hline
AHMR (Ours) & {\bf 0.20} & 0.50 & {\bf 0.86} & {\bf 1.08} & {\bf 0.40} & {\bf 0.72} & {\bf 1.03} & {\bf 1.23} & {\bf 0.14} & {\bf 0.32} & {\bf 0.50} & {\bf 0.54} & {\bf 0.25} & {\bf 0.50} & {\bf 0.79} & {\bf 0.91} \\ \hline
\end{tabular}
}
\caption{\footnotesize{Performance evaluation (in MAE) of comparison methods over all action types on the H3.6m dataset.}}
\label{tab:h3.6m_mae}
\end{table*}

\begin{table*}[ht]
\centering
\resizebox{\textwidth}{!}{
\renewcommand\tabcolsep{6.0pt}
\begin{tabular}{|c|ccccc|ccccc|ccccc|ccccc|}
\hline
Time (ms) & 80 & 160 & 320 & 400 & 1,000 & 80 & 160 & 320 & 400 & 1,000 & 80 & 160 & 320 & 400 & 1,000 & 80 & 160 & 320 & 400 & 1,000 \\ \hline\hline

& \multicolumn{5}{c|}{Basketball}  & \multicolumn{5}{c|}{Basketball Signal}    & \multicolumn{5}{c|}{Directing Traffic}         & \multicolumn{5}{c|}{Jumping}      \\ \hline

ResGRU~\cite{residualgru} & 0.49 & 0.77 & 1.26 & 1.45 & 1.77 & 0.42 & 0.76 & 1.33 & 1.54 & 2.17 & 0.31 & 0.58 & 0.94 & 1.10 & 2.06 & 0.57 & 0.86 & 1.76 & 2.03 & 2.42 \\ \hline

CEM~\cite{li2018convolutional} & 0.36 & 0.62 & 1.07 & 1.17 & 1.95 & 0.33 & 0.62 & 1.05 & 1.23 & 1.98 & 0.26 & 0.58 & 0.91 & 1.04 & 2.08 & 0.38 & 0.60 & 1.36 & 1.58 & 2.05 \\ \hline

Traj-GCN~\cite{mao2019learning} & 0.33 & 0.52 & 0.89 & 1.06 & 1.71 & 0.11 & 0.20 & 0.41 & 0.53 & 1.00 & 0.15 & 0.32 & 0.52 & 0.60 & 2.00 & 0.31 & 0.49 & 1.23 & 1.39 & 1.80 \\ \hline

DMGNN~\cite{li2020dynamic} & 0.30 & 0.46 & 0.89 & 1.11 & 1.66 & 0.10 & 0.17 & 0.31 & 0.41 & 1.26 & 0.15 & 0.30 & 0.57 & 0.72 & 1.98 & 0.37 & 0.65 & 1.49 & 1.71 & 1.79 \\ \hline

AHMR(Ours) & \textbf{0.19} & \textbf{0.38}& \textbf{0.69} & \textbf{0.84} & \textbf{1.62} 
& \textbf{0.06} & \textbf{0.12}& \textbf{0.24} & \textbf{0.29} & \textbf{0.76}
& \textbf{0.12} & \textbf{0.23} & \textbf{0.43} & \textbf{0.54} & \textbf{1.24}
& \textbf{0.19} & \textbf{0.32} & \textbf{0.57} & \textbf{0.67} & \textbf{1.00} \\ \hline\hline

& \multicolumn{5}{c|}{Running}  & \multicolumn{5}{c|}{Soccer}     & \multicolumn{5}{c|}{Walking}    & \multicolumn{5}{c|}{Washing Window} \\ \hline

ResGRU~\cite{residualgru} & 0.32 & 0.48 & 0.65 & 0.74 & 1.00 & 0.29 & 0.50 & 0.87 & 0.98 & 1.73 & 0.35 & 0.45 & 0.59 & 0.64 & 0.88 & 0.31 & 0.47 & 0.74 & 0.93 & 1.37 \\ \hline

CEM~\cite{li2018convolutional} & 0.28 & 0.43 & 0.54 & 0.57 & 0.69 & 0.28 & 0.48 & 0.79 & 0.90 & 1.58 & 0.35 & 0.44 & 0.46 & 0.51 & 0.77 & 0.30 & 0.47 & 0.79 & 1.00 & 1.39 \\ \hline

Traj-GCN~\cite{mao2019learning} & 0.33 & 0.55 & 0.73 & 0.74 & 0.95 & 0.18 & 0.29 & 0.61 & 0.71 & 1.40 & 0.33 & 0.45 & 0.49 & 0.53 & 0.61 & 0.22 & 0.33 & 0.57 & 0.75 & 1.20 \\ \hline

DMGNN~\cite{li2020dynamic} & 0.19 & 0.31 & 0.47 & \textbf{0.49} & \textbf{0.64} & 0.22 & 0.32 & 0.79 & 0.91 & 1.54 & 0.30 & 0.34 & 0.38 & 0.43 & 0.60 & 0.20 & 0.27 & 0.62 & 0.81 & 1.09 \\ \hline

AHMR(Ours) & \textbf{0.18} & \textbf{0.30} & \textbf{0.46} & 0.50 & 0.85
& \textbf{0.21} & \textbf{0.34} & \textbf{0.54} & \textbf{0.66} & \textbf{1.39} 
& \textbf{0.16} & \textbf{0.30} & \textbf{0.36} & \textbf{0.40} & \textbf{0.55} 
& \textbf{0.14} & \textbf{0.22} & \textbf{0.35} & \textbf{0.41} & \textbf{0.86} \\ \hline
\end{tabular}}
\caption{\footnotesize{Performance evaluation (in MAE) of comparison methods over 8 action types on the CMU MoCap dataset.}}
\label{tab:cmu}
\end{table*}

\section{Experiments}
In this section, we conduct extensive experiments on three large and complex datasets of distinct articulate objects, \emph{namely human, fish, and mouse}, to evaluate the proposed method. We aim to answer the following research questions:
\begin{itemize}
\setlength{\leftmargin}{10em}
\setlength{\listparindent}{10em}
\item \textbf{RQ1:} How does the proposed method compare against the state-of-the-art methods for human motion prediction?
\item \textbf{RQ2:} How do the different pose representations and loss functions compare against each other empirically?
\item \textbf{RQ3:} How much do different components of \emph{AHMR} contribute to the performance? What is the efficiency of \emph{AHMR} compared to existing methods?
\item \textbf{RQ4:} How well does the proposed method generalise to the fish and mouse datasets?
\end{itemize}

We first introduce the experimental settings, then proceed to answer the above research questions.

\subsection{Experimental Settings}
\textbf{Datasets} \quad For \emph{human motion prediction}, we engage two widely employed benchmark datasets, H3.6m and CMU MoCap. H3.6m~\cite{H36M} is the largest public full-body motion dataset, containing 3.6 million 3D human poses with 15 activities performed by 7 subjects. For the CMU MoCap dataset~\cite{cmu}, experiments are performed on 8 action categories. We follow the same training/test dataset split and training protocols as previous works~\cite{srnn,residualgru}. For \emph{fish motion prediction}, we consider the fish dataset of~\cite{Liex}, which contain 14 fish videos (50 FPS) of 6 different fishes. In general, lengths of the continuous sequences in these videos vary from 2,250 frames to 24,000 frames. For \emph{mouse motion prediction}, we construct a 3D mouse everyday motion dataset. Mouse depth images are acquired with a top-view Primesense Carmine camera at 25 FPS. The mouse 3D poses are extracted from depth images using our annotation tool. Different from the H3.6M dataset that cater for short motion sequences, lengths of the continuous sequences for mouse in our datasets are significantly longer, which may extend to 30,000 frames (10 minutes). In total, there are 12 videos of 4 mice.

\textbf{Parameter Settings} \quad The hidden state size, \emph{i.e.}, dimension of state vectors ${\mathbf{s}}$ and ${\mathbf{g}}$ is set to 500, 200, and 100 respectively for human, fish, and mouse motion prediction. All other settings and hyperparameters are constant across different objects. The default number of recurrent steps is set to 5 and neighboring context window size 3. Following previous works~\cite{srnn, residualgru}, we ignore global translation and utilize $t=50$ observed frames as inputs to predict future $T=10$ frames in training. The Adam optimizer is employed with a learning rate initialized as 0.001 and decaying by 0.95 every 5,000 iterations. A batch size of 16 is used and the gradient clipping threshold is set to 5. Typically, training converges in 30,000 iterations on the H3.6m dataset, 15,000 iterations on the fish dataset and 10,000 iterations on the mouse dataset.

\subsection{Comparison with Existing Methods (RQ1)} \label{sec:quantitative}
\begin{figure*}[ht!]
\begin{center}
\includegraphics[width=0.95\textwidth]{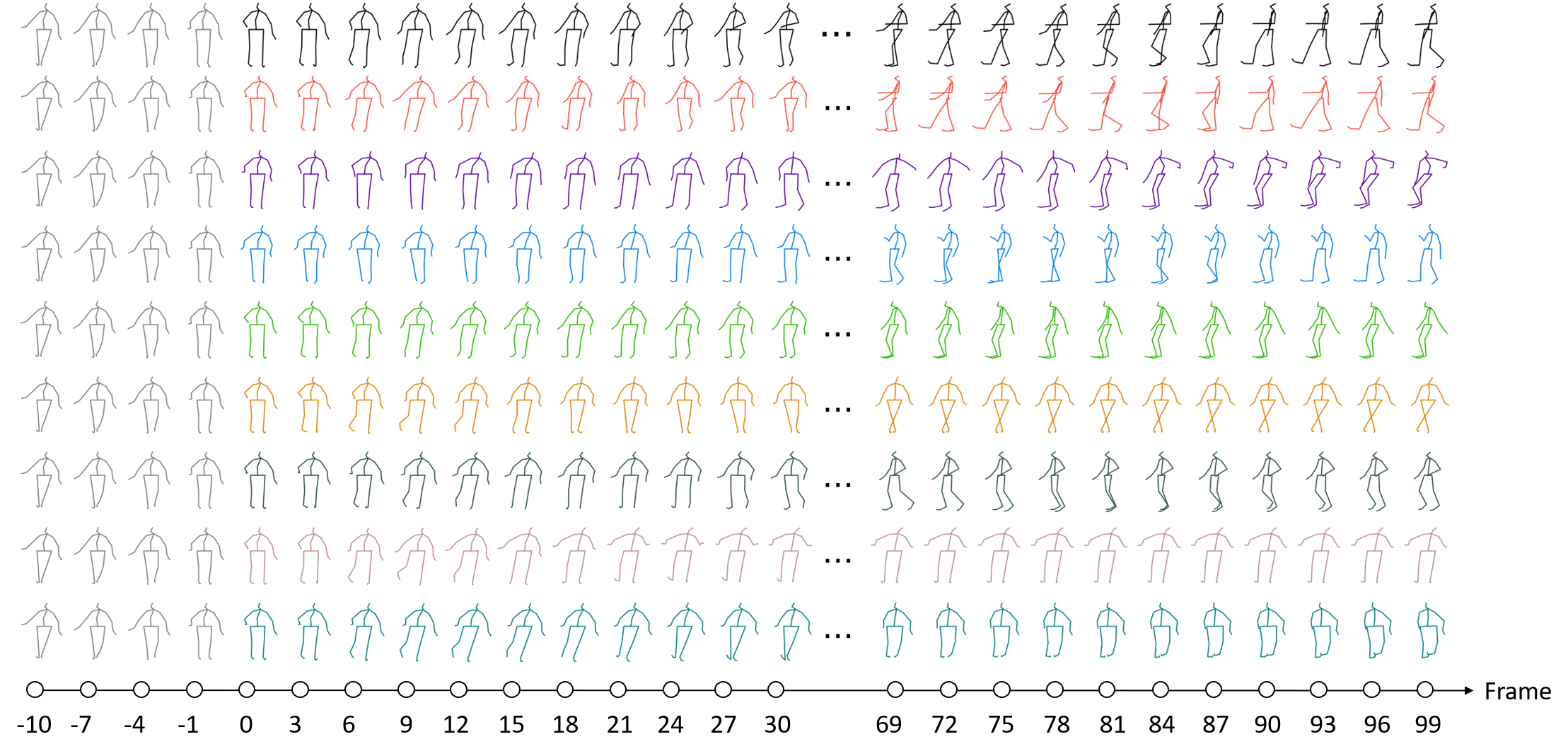}
\caption{\footnotesize{Motion forecasting of walking activity by the comparison methods on the H3.6m dataset. 1st line: the ground truth; 2nd line: our method; Existing methods are shown on the 3rd to 8th line: ERD~\cite{erd}, SRNN~\cite{srnn}, ResGRU~\cite{residualgru}, QuaterNet~\cite{pavllo2019modeling}, CEM~\cite{li2018convolutional} \& DMGNN~\cite{li2020dynamic}. The last line shows a baseline where 3D coordinates representation is employed. The complete visual results can be found in the supplementary video file.}}
\label{fig:walking}
\end{center}
\end{figure*}
\label{exp_human}

\subsubsection{Short-term evaluation in the MAE metric}
We first benchmark the state-of-the-art motion-prediction methods on the H3.6m dataset~\cite{H36M}, employing the commonly adopted mean angle error (MAE) metric~\cite{srnn, residualgru}. More specifically, we denote $\alpha_i,\beta_i,\gamma_i$ as the Euler angles between $i^{th}$ bone and $i-1^{th}$ bone. For a skeleton of $m$ bones, the MAE is evaluated as:
\begin{equation}
\resizebox{.9\linewidth}{!}{$\begin{aligned}
\textrm{MAE}= \frac{1}{m-1}\sum_{i=2}^{m} \norm{(\alpha_i-\hat{\alpha}_i) + (\beta_i-\hat{\beta}_i) + (\gamma_i-\hat{\gamma}_i)}_{2},\end{aligned}$}
\end{equation}
namely the difference of ground truth and predicted Euler angles. Note that as per the standard evaluation protocol, evaluating the MAE does not take into account the rotation angles of the root joint, \emph{i.e.}, the global rotation.

The performance of different methods are summarized in terms of MAE in Table~\ref{tab:h3.6m_mae}. Our \emph{AHMR} employs the Stiefel manifold for pose parameterization, and is trained with the geodesic loss function. In total, ten methods are compared, including ERD~\cite{erd}, SRNN~\cite{srnn}, Res-GRU~\cite{residualgru}, QuaterNet~\cite{pavllo2019modeling}, CEM~\cite{li2018convolutional}, AGED~\cite{AGED}, Traj-GCN~\cite{mao2019learning}, DMGNN~\cite{li2020dynamic}, Transformer~\cite{aksan2020spatio}, and our method. Zero-Velocity is a baseline that simply outputs the last input frame as the prediction. ERD~\cite{erd} incorporates nonlinear encoder and decoder layers on top of recurrent networks. SRNN~\cite{srnn} represents the human body with a spatio-temporal graph which is fed into a LSTM architecture.
Res-GRU~\cite{residualgru} adopts GRU to forecast first order differences in the axis angle representation whereas QuaterNet~\cite{pavllo2019modeling} utilises a quaternion representation. AGED~\cite{AGED} incorporates adversarial training with two discriminators on top of the RNN encoder and decoder modules. CEM~\cite{li2018convolutional} utilises a convolutional network,~\cite{aksan2020spatio} uses a Transformer network, whereas Traj-GCN~\cite{mao2019learning} and DMGNN~\cite{li2020dynamic} both employ graph neural network for encoding the motion sequence inputs. We reproduce the existing methods following their released codes on GitHub where possible. For fair comparison, we followed the settings of prior works~\cite{erd,srnn,residualgru}: 
1) In short-term ($\leqslant$ $T=10$ frames or 400 ms) prediction, training is done over all activity types with a input window size of 50 frames and a training output window size of $T=10$ frames; 
2) Subject 5 is used for testing while the rest are used for training.

For \emph{short-term} ($\leqslant 400ms$) motion prediction, from the empirical evidences in Table~\ref{tab:h3.6m_mae}, we observe that our proposed \emph{AHMR} delivers superior performance over the state-of-the-art methods. Interestingly, a couple of earlier approaches including ERD~\cite{erd} and SRNN~\cite{srnn} are outperformed by the Zero-Velocity baseline, which simply replicates the last observed pose as future pose predictions. These quantitative results reveal that existing methods suffer from the issue of a clear inaccurate short-term predictions. Subsequent methods seek to address this issue by modeling (angular) velocities~\cite{residualgru, pavllo2019modeling} and incorporating temporal smoothing. These temporal smoothing schemes include adversarial approaches~\cite{AGED}, learning local temporal dependencies via convolutional layers~\cite{li2018convolutional, li2020dynamic} or discrete Fourier transforms~\cite{mao2019learning}. Our method models the entire observed sequence globally and simultaneously with attention. This gives our approach an edge over the local temporal modeling in convolutional based approaches, which translates to better quantitative results in the short term.

We further benchmark our method on the CMU MoCap dataset and report the MAE evaluation results in Table~\ref{tab:cmu}. The 8 activity categories contain both periodic and non-periodic actions. The quantitative results indicate that AHMR clearly outperforms the state-of-the-art methods across all actions. This provides further evidence for the robustness and effectiveness of the proposed method.

\begin{table*}[ht!]
\resizebox{1\textwidth}{!}
{
\begin{tabular}{|c|cccc|cccc|cccc|cccc|}
\hline
Time(ms) & 80 & 160 & 320 & 400 & 80 & 160 & 320 & 400 & 80 & 160 & 320 & 400 & 80 & 160 & 320 & 400 \\ \hline\hline

& \multicolumn{4}{c|}{Walking} & \multicolumn{4}{c|}{Eating} & \multicolumn{4}{c|}{Smoking} & \multicolumn{4}{c|}{Discussion} \\ \hline
ResGRU~\cite{residualgru} & 20.5 & 39.8 & 78.2 & 90.3 & 17.5 & 34.3 & 71.1 & 87.5 & 22.4 & 39.9 & 80.2 & 92.5 & 25.8 & 43.4 & 83.5 & 95.8 \\ \hline
CEM~\cite{li2018convolutional} & 17.1 & 31.2 & 53.8 & 61.5 & 13.7 & 25.9 & 52.5 & 63.3 & 11.1 & 21.0 & 33.4 & 38.3 & 18.9 & 39.3 & 67.7 & 75.7 \\ \hline
Traj-GCN~\cite{mao2019learning} & 8.9 & 15.7 & 29.2 & 33.4 & 8.8 & 18.9 & 39.4 & 47.2 & 7.8 & 14.9 & 25.3 & 28.7 & 9.8 & 22.1 & 39.6 & 39.9 \\ \hline
DMGNN~\cite{li2020dynamic} & 8.9 & 14.9 & 29.0 & 33.1 & 8.7 & 18.7 & 39.5 & 47.1 & 8.2 & 14.5 & 25.1 & 58.8 & 9.7 & 21.9 & 39.5 & 40.0 \\ \hline
AHMR (Ours) & \textbf{7.8} & \textbf{12.2} & \textbf{16.2} & \textbf{19.7} & \textbf{4.9} & \textbf{8.4} & \textbf{14.0} & \textbf{17.4} & \textbf{3.6} & \textbf{6.5} & \textbf{11.2} & \textbf{13.1} & \textbf{5.8} & \textbf{10.6} & \textbf{16.1} & \textbf{17.0} \\ \hline\hline

& \multicolumn{4}{c|}{Directions} & \multicolumn{4}{c|}{Greeting} & \multicolumn{4}{c|}{Phoning} & \multicolumn{4}{c|}{Posing} \\ \hline
ResGRU~\cite{residualgru} & 36.4 & 56.6 & 80.3 & 98.1 & 36.8 & 73.3 & 138.2 & 155.6 & 24.3 & 42.3 & 72.6 & 82.3 & 26.7 & 52.4 & 129.5 & 159.4 \\ \hline
CEM~\cite{li2018convolutional} & 22.0 & 37.2 & 59.6 & 73.4 & 24.5 & 46.2 & 90.0 & 103.1 & 17.2 & 29.7 & 53.4 & 61.3 & 16.1 & 35.6 & 86.2 & 105.6 \\ \hline
Traj-GCN~\cite{mao2019learning} & 12.6 & 24.4 & 48.2 & 58.4 & 14.5 & 30.5 & 74.2 & 89.0 & 10.4 & 14.3 & 33.1 & 39.7 & 9.4 & 23.9 & 66.2 & 82.9 \\ \hline
DMGNN~\cite{li2020dynamic} & 12.3 & 23.8 & 46.2 & 55.5 & 14.0 & 29.8 & 74.0 & 140.2 & 10.2 & 14.0 & 32.8 & 40.0 & 9.2 & 23.5 & 65.0 & 82.8 \\ \hline
AHMR (Ours) & \textbf{7.3} & \textbf{11.4} & \textbf{15.2} & \textbf{18.7} & \textbf{7.1} & \textbf{13.5} & \textbf{25.9} & \textbf{29.9} & \textbf{6.6} & \textbf{11.1} & \textbf{18.3} & \textbf{20.7} & \textbf{5.5} & \textbf{10.9} & \textbf{25.0} & \textbf{31.2} \\ \hline\hline

& \multicolumn{4}{c|}{Purchases} & \multicolumn{4}{c|}{Sitting} & \multicolumn{4}{c|}{Sitting Down} & \multicolumn{4}{c|}{Taking Photo} \\ \hline 
ResGRU~\cite{residualgru} & 38.5 & 70.1 & 101.0 & 102.3 & 34.1 & 53.2 & 110.4 & 115.0 & 28.6 & 55.2 & 85.6 & 115.8 & 23.1 & 47.0 & 92.3 & 110.1 \\ \hline
CEM~\cite{li2018convolutional} & 29.4 & 54.9 & 82.2 & 93.0 & 19.8 & 42.4 & 77.0 & 88.4 & 17.1 & 34.9 & 66.3 & 77.6 & 14.0 & 27.2 & 53.8 & 66.2 \\ \hline
Traj-GCN~\cite{mao2019learning} & 19.6 & 38.5 & 64.4 & 72.2 & 10.7 & 24.6 & 50.6 & 62.0 & 11.4 & 27.6 & 56.4 & 67.6 & 6.8 & 15.2 & 38.2 & 49.6 \\ \hline
DMGNN~\cite{li2020dynamic} & 19.3 & 38.0 & 64.2 & 72.1 & 10.6 & 24.4 & 50.3 & 61.8 & 11.2 & 27.5 & 56.1 & 67.7 & 7.1 & 15.0 & 38.1 & 49.5 \\ \hline
AHMR (Ours) & \textbf{7.2} & \textbf{12.8} & \textbf{19.6} & \textbf{22.8} & \textbf{7.0} & \textbf{13.9} & \textbf{24.6} & \textbf{27.4} & \textbf{5.5} & \textbf{10.2} & \textbf{19.8} & \textbf{23.4} & \textbf{4.4} & \textbf{8.2} & \textbf{15.6} & \textbf{18.4} \\ \hline\hline

& \multicolumn{4}{c|}{Waiting} & \multicolumn{4}{c|}{Walking Dog} & \multicolumn{4}{c|}{Walking Together} & \multicolumn{4}{c|}{Average} \\ \hline
ResGRU~\cite{residualgru} & 29.5 & 60.4 & 118.1 & 138.5 & 59.8 & 78.6 & 152.3 & 178.3 & 25.4 & 53.2 & 89.8 & 99.6 & 30.0 & 53.3 & 98.9 & 114.7 \\ \hline
CEM~\cite{li2018convolutional} & 17.9 & 36.5 & 74.9 & 90.7 & 40.6 & 74.7 & 116.6 & 138.7 & 15.0 & 29.9 & 54.3 & 65.8 & 19.6 & 37.8 & 68.1 & 80.2 \\ \hline
Traj-GCN~\cite{mao2019learning} & 9.5 & 22.0 & 57.5 & 73.9 & 32.2 & 58.0 & 102.2 & 122.7 & 8.9 & 18.4 & 35.3 & 44.3 & 12.1 & 24.6 & 50.7 & 60.8 \\ \hline
DMGNN~\cite{li2020dynamic} & 9.6 & 21.8 & 56.9 & 71.9 & 31.8 & 58.3 & 101.9 & 122.4 & 8.8 & 18.0 & 35.5 & 44.2 & 12.0 & 24.3 & 50.3 & 62.4 \\ \hline
AHMR (Ours) & \textbf{6.7} & \textbf{13.7} & \textbf{27.8} & \textbf{33.0} & \textbf{12.4} & \textbf{19.1} & \textbf{28.5} & \textbf{35.2} & \textbf{5.6} & \textbf{10.3} & \textbf{16.2} & \textbf{18.9} & \textbf{6.5} & \textbf{11.5} & \textbf{19.6} & \textbf{23.1} \\ \hline
\end{tabular}
}
\caption{\footnotesize{Performance evaluation (in MPE) of comparison methods over all action types on the H3.6m dataset.}}
\label{tab:h3.6m_mpe}
\end{table*}

\subsubsection{Short-term evaluation in the MPE metric}
In some applications, predicting the positions of the skeletal joints would be more useful than the orientation angles of the bones. As such, we also evaluate our performance on the mean position error (MPE) metric. We align global translation but not global rotation. With $\bm{J}_i$ denoting the 3D position of joint $i$, the MPE is defined as:
\begin{equation}
\begin{aligned}
\textrm{MPE}= \frac{1}{m}\sum_{i=1}^{m} \norm{\bm{J}_i - \hat{\bm{J}}_i}_2.
\end{aligned}
\end{equation}

Our \emph{AHMR} still employs the Stiefel manifold parameterization, but is trained with the forward kinematics loss for this setting. The evaluation results for the MPE metric on the H3.6m dataset are reported in Table~\ref{tab:h3.6m_mpe}. Our method consistently surpass the current state-of-the-arts by a significant margin. In large part, this could be attributed to the fact that the forward kinematics loss is more suitable when the evaluation metric is the MPE. The forward kinematics loss respects the hierarchical nature of kinematic chains, and reduces error accumulation traversing down the chains.

\subsubsection{Long-term and qualitative evaluation}
Whereas accurate prediction is possible and desirable for the short-term, it is unrealistic to expect accurate forecasting in the long-term simply due to the stochastic nature of motion tendencies in different subjects. A more reasonable goal is to achieve natural looking human-like motions that demonstrate consistency and faithfulness. We therefore perform visual evaluation for long-term prediction. Figure~\ref{fig:walking} illustrates the visual results of the walking activity for a period of 100 frames (4 seconds). More results are presented in the supplementary video. We note that~\cite{mao2019learning} employs discrete cosine transforms to map the temporal inputs into the frequency domain, and fixes the output sequence to a fixed number of frames. Consequently, forecasting for the arbitrary long-term future is not possible in~\cite{mao2019learning}.

From the figure and the supplementary file, we make the observations that in the relatively short term ($<$25 frames or 1 second), existing methods are capable of predicting human like motion. However, most existing methods have the issue of converging to a motionless mean pose or displaying unrealistic motion in the longer term. For example, ERD~\cite{erd} drifts away to non-human like motion while SRNN~\cite{srnn}, Res-GRU~\cite{residualgru}, QuaterNet~\cite{pavllo2019modeling} and DMGNN~\cite{li2020dynamic} converge to a motionless pose. In contrast, the proposed method \emph{AHMR} is capable of producing human-like motion predictions even after 1,000 frames. In this regard, an important highlight of our architecture is the capability to generate natural long-term motion predictions.

\subsection{Comparison of Different Pose Representations and Loss Functions (RQ2)} \label{sec:poserep&loss}
\begin{table*}[ht!]
\centering
\resizebox{1\textwidth}{!}
{
\begin{tabular}{|c|cccc|cccc|c|c|c|c|}
\hline
\multirow{2}{*}{Evaluation Metric} & \multicolumn{4}{c|}{Representation} & \multicolumn{4}{c|}{Loss Function} & \multicolumn{4}{c|}{Time (ms)} \\ \cline{2-13} 
& 3D Coordinates & Axis Angles & Quaternions & Stiefel Manifold & L2 & Smooth L1 & Geodesic & Forward Kinematics & 80 & 160 & 320 & 400 \\ \hline\hline
\multirow{10}{*}{MAE}
& \checkmark &  &  &  & \checkmark &  &  &  & 0.32 & 0.8 & 1.09 & 1.25 \\ \cline{2-13} 
&  & \checkmark &  &  & \checkmark &  &  &  & 0.29 & 0.55 & 0.89 & 1.04 \\ \cline{2-13} 
&  &  & \checkmark &  & \checkmark &  &  &  & 0.30 & 0.57 & 0.88 & 1.02 \\ \cline{2-13} 
&  &  &  & \checkmark & \checkmark &  &  &  & 0.27 & 0.53 & 0.85 & 0.99 \\ \cline{2-13} 
&  & \checkmark &  &  &  & \checkmark &  &  & 0.31 & 0.58 & 0.92 & 1.05 \\ \cline{2-13} 
&  &  & \checkmark &  &  & \checkmark &  &  & 0.31 & 0.62 & 0.91 & 1.09 \\ \cline{2-13} 
&  &  &  & \checkmark &  & \checkmark &  &  & 0.29 & 0.56 & 0.83 & 1.01 \\ \cline{2-13} 
&  & \checkmark &  &  &  &  & \checkmark &  & 0.32 & 0.58 & 0.91 & 1.09 \\ \cline{2-13} 
&  &  & \checkmark &  &  &  & \checkmark &  & 0.31 & 0.57 & 0.93 & 1.10 \\ \cline{2-13} 
&  &  &  & \checkmark &  &  & \checkmark &  & \textbf{0.25} & \textbf{0.50} & \textbf{0.79} & \textbf{0.91} \\ \hline\hline
\multirow{6}{*}{MPE} &   & \checkmark  &  &  &  &  &  & \checkmark  & \multicolumn{4}{c|}{Fails to Converge} \\ \cline{2-13} 
&  &   & \checkmark &  &  &  &  &  \checkmark & \multicolumn{4}{c|}{Fails to Converge}  \\ \cline{2-13}
&  &  &  & \checkmark & \checkmark &  &  &  & 13.4 & 27.6 & 57.6 & 78.1 \\ \cline{2-13} 
&  &  &  & \checkmark &  & \checkmark &  &  & 14.3 & 29.5 & 56.8 & 73.5 \\ \cline{2-13} 
&  &  &  & \checkmark &  &  & \checkmark &  & 11.8 & 20.6 & 46.2 & 58.4 \\ \cline{2-13} 
&  &  &  &  \checkmark &  &  &  & \checkmark & \textbf{6.5} & \textbf{11.5} & \textbf{19.6} & \textbf{23.1} \\ \hline
\end{tabular}
}
\caption{\footnotesize{Comparison between different pose representations and loss functions.}} \label{tab:representations}
\end{table*}

\textbf{Qualitative Comparison} \quad
As shown in Figure 6 and the supplementary file, the 3D coordinates representations resulted in varying bone lengths and body distortions. We could alleviate the varying bone lengths issue via an explicit bone length preserving loss term, but the body distortion issues such as zigzag predictions for the central torso remain. This highlights severe limitations of raw coordinates representation and the advantage of incorporating a kinematic tree representation which naturally preserves bone lengths and body prior.

We further perform quantitative evaluations on various skeletal representations and different loss functions. We tried 16 different combinations of the skeletal representations and the loss functions. The results averaged over all 15 activities in the H3.6m dataset are reported in Table 4. For a fair comparison, all the different combinations are coupled with the same \emph{AHMR} network.

\textbf{Insights for Quantitative Comparison in MAE.} \quad
For the quantitative comparison on the commonly adopted MAE metric, we have the following insights.
\begin{enumerate}
\item In general, the 3D coordinates representation is shown to have consistently lower performance than other representations.
\item For axis angle and quaternion representations, L2 loss is more suitable to be coupled with them. The geodesic loss does not pair well with the axis angle or quaternion representations and in fact result in slight performance dips compared to the L2 loss. This is in line with our analysis in Section 3.2  and can be understood from the additional computational complexity incurred by converting the axis angles or quaternions to rotation matrices for computing the geodesic loss. Such computations are not optimized for backpropagations and prone to gradient vanishing especially when the involved quantities are of small magnitudes.
\item The combination of Stiefel manifold with the geodesic loss consistently delivers the best performance, with pronounced improvements over the L2 loss and smooth L1 loss. Obtaining rotation matrices from Stiefel manifolds involves a very efficient cross product computation, facilitating the computation of the geodesic loss, thus making this combination an ideal pair.
\item The Stiefel manifold clearly outperforms the axis angle and quaternion representations for various loss functions.
\end{enumerate}

\textbf{Insights for Quantitative Comparison in MPE.} \quad
We now evaluate the performance of various loss functions with respect to the MPE metric. Notably, the forward kinematics loss fails to converge for the axis angle and quaternion representations. This can be understood as resulting from the computational burden of obtaining the rotation matrices from these parameterizations. Consequently, we centre in onto the Stiefel manifold representation. We make the following key observations.
\begin{enumerate}
\item The forward kinematics loss convincingly outperforms the other loss functions with a more than 44\% quantitative improvement. This is consistent with our theoretical expectations that the forward kinematics loss and the MPE metric are aligned. The forward kinematics loss takes into account the higher significance of rotations nearer to the chain root. It is thus more suitable when we are interested in benchmarking our performance via a mean position error metric (MPE). The geodesic loss or L2 loss instead assigns an equal weight to each bone orientation, and are thus more suitable for the mean angle error (MAE) metric.
\item Geodesic loss still outperforms the conventional L2 loss in MPE when coupled with the Stiefel Manifold representation. 
\item Both the axis angle and the quaternion representations suffer from non-convergence problems when combined with the forward kinematics loss. This mainly dues to the computation of rotation matrices, which, however, is not an issue for the Stiefel manifold representation. 
\end{enumerate}

\subsection{Studies on Key Components of \emph{AHMR} (RQ3)} \label{sec:ahmrsettings}
\textbf{AHMR settings}
\begin{table}[ht!]
\centering
\resizebox{1\linewidth}{!}{
\begin{tabular}{|c|c|c||c|c|c||c|c|c|}
\hline
Hidden Size & 80 ms & 400 ms & Rec. Steps & 80 ms & 400 ms & Neigh. Win. & 80 ms & 400 ms \\ \hline
128 & 0.30 & 0.98 & 1 & 0.29 & 0.96 & 1 & 0.29 & 0.96 \\\hline
\textbf{256} & 0.25 & 0.91 & 3 & 0.27 & 0.94 & \textbf{3} & 0.25 & 0.91 \\\hline
512 & 0.27 & 0.96 & \textbf{5} & 0.25 & 0.91 & 5 & 0.26 & 0.94 \\\hline
1024 & 0.28 & 0.95 & 7 & 0.26 & 0.94 & 7 & 0.27 & 0.93 \\\hline
\end{tabular}
}
\caption{\footnotesize{MAE averaged over all activities on H3.6m, obtained by varying the internal parameters, including dimension of hidden states, number of recurrent steps $n$, and context neighboring window size. Boldface denotes the default values.}}
\label{tab:AHMRsettings}
\end{table}

We conduct experiments to determine the optimal settings for \emph{AHMR}. This includes the dimension of hidden units, number of recurrent steps and neighboring context window size. At each recurrent step, \emph{AHMR} allows information exchange of a local state within its context window. It is interesting to see the effect of enlarging or reducing the neighboring window size and the number of recurrent steps. As shown in Table~\ref{tab:AHMRsettings}, we found that enlarging the neighboring context window size does not necessarily leads to significant improvement in accuracy. By exploring the hyperparameter space empirically and considering the parameter size, we settled on the optimal values of 256 for the hidden units dimensions, 5 for the number of recurrent steps and 3 for the context window size.

\textbf{Further Ablation Studies on \emph{AHMR} Architecture}
\begin{table}[ht!]
\resizebox{1\linewidth}{!}{
\begin{tabular}{|l|c|c|c|c|c|}
\hline
& \multicolumn{5}{c|}{Average} \\ \hline
Time(ms) & 80 & 160 & 320 & 400 & 1000 \\ \hline
Remove left \& right forget gates & 0.30 & 0.56 & 0.88 & 1.03 & 1.66 \\ \hline
Remove sequence level state & 0.28 & 0.53 & 0.84 & 0.96 & 1.73 \\ \hline
Remove spatial attention & 0.29 & 0.54 & 0.85 & 0.95 & 1.65 \\ \hline
Remove spatial \& temporal attention & 0.31 & 0.56 & 0.89 & 1.05 & 1.74 \\ \hline
Single decoder layer & 0.26 & 0.53 & 0.85 & 0.97 & 1.62 \\ \hline
Decoder layer for each recurrent state & 0.27 & 0.55 & 0.84 & 0.94 & 1.61 \\ \hline
AHMR (Full) & \textbf{0.25} & \textbf{0.50} & \textbf{0.79}
& \textbf{0.91} & \textbf{1.58} \\ \hline
\end{tabular}
}
\caption{\footnotesize{Performance evaluation (in MAE) of variants of our \emph{AHMR} network averaged over all activities on the H3.6m dataset.}}
\label{tab:ablation}
\end{table}

We further investigated several variants of our \emph{AHMR} network: 
\begin{enumerate}
\item Removal of the left and right forget gates in the update of the frame level states ${\mathbf{h}}_j$ from the encoder.
\item Removal of the sequence level state $\textbf{g}$ from the encoder.
\item Removal of the spatial attention.
\item Removal of both spatial and temporal attention.
\item Using a single layer for decoder network.
\item Setting the number of layers in decoder to that of recurrent steps $n$, \emph{i.e.}, each recurrent state is fed to a decoder layer.
\end{enumerate}

As shown in Table~\ref{tab:ablation}, prediction accuracy suffered upon removal of the left and right forget gates, which is especially true for the shorter term (up to 400ms). On the other hand, removal of the sequence level state $\textbf{g}$ from our \emph{AHMR} encoder is observed not to severely affect forecasting before 400 ms but performance beyond 400 ms declines noticeably. We may rationalize the sequence level state as being more effective for capturing long term dependencies, which translates to long term prediction capabilities.

The effectiveness of the spatial and temporal attention mechanisms are illustrated when their removals result in noticeable performance deterioration. In particular, it may be seen that among these ablation experiments, removal of the spatio-temporal attention totally resulted in the most significant performance drop.

Interestingly, changing the decoder architecture does not affect the prediction accuracy as much. This suggests that the encoder, which is responsible for context modeling, is more important than the decoder, which is responsible for translating the motion context in an autoregressive fashion.

\textbf{Computational efficiency}
\begin{table}[ht!]
\centering
\resizebox{1\linewidth}{!}
{
\begin{tabular}{|l|c|c|c|}
\hline
Methods & \# Parameters & Train time / 1,000 iterations (s) & Test FPS\\ \hline
ERD & 17,348,054 & 428 & 52\\ \hline
SRNN & 22,817,888 & 947 & 15\\ \hline
Res-GRU & 6,684,726 & 80 & 173\\ \hline
AHMR (Ours) & \textbf{3,233,334} & \textbf{24} & \textbf{636}\\ \hline
\end{tabular}
}
\caption{\footnotesize{Number of training parameters and efficiency.}}
\label{tab:efficiency}
\end{table}

We further evaluate the efficiency of \emph{AHMR} against existing methods. The training and testing time as well as number of training parameters required for different methods are shown in Table~\ref{tab:efficiency}. Our architecture was implemented using TensorFlow 1.14. All experiments were performed on a Nvidia GeForce GTX TITAN X GPU. From the table, we can see that \emph{AHMR} requires less parameters than existing methods and its computation speed is significantly faster. In particular, \emph{AHMR} achieves 636 FPS (frames per second) in testing, while the state-of-the-art is 173 FPS. We conjecture that the superior efficiency performance mainly stems from the shallower and non-sequential network structure of \emph{AHMR} comparing to  the state-of-the-art methods.
\begin{figure}[ht!]
\centering
\includegraphics[width=0.95\linewidth]{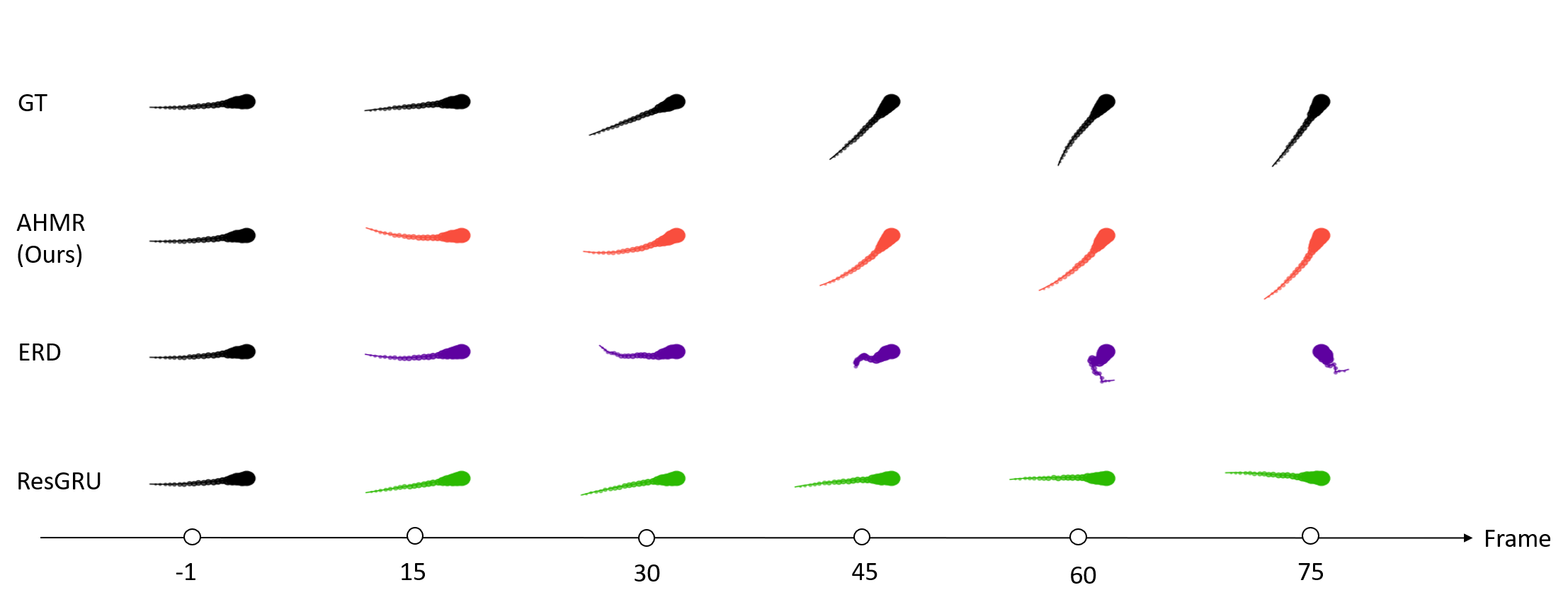}
\caption{\footnotesize{Motion forecasting on Fish dataset. The head of the fish is rendered wider for resemblance with the actual zebrafish.}}
\label{fig:Fish}
\end{figure}
\begin{figure}[ht!]
\centering
\includegraphics[width=0.95\linewidth]{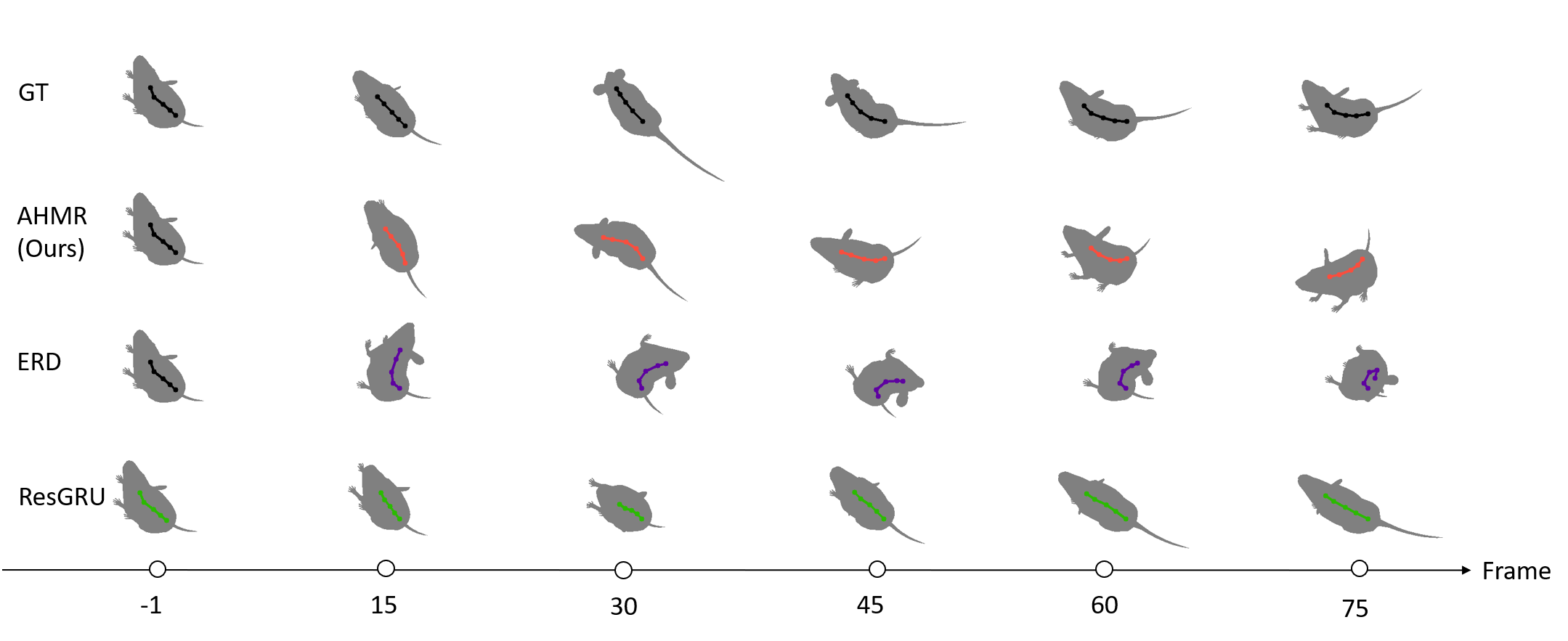}
\caption{\footnotesize{Motion forecasting on Mouse dataset. The mouse shape is rendered in gray color with joints marked out along the spine.}}
\label{fig:Mouse}
\end{figure}

\subsection{Performance Comparison on Fish and Mouse Motion Prediction (RQ4)}
\label{exp_animal}
In this subsection, we evaluate how our method generalize and perform on on the fish and mouse datasets. Both the fish and mouse are represented by a single kinematic chain and are simpler than human, which is represented as five kinematic chains. Consequently, instead of ignoring the global rotation as was the case for H3.6m, we decided to include it into the MAE evaluation since it is reasonable that any effective prediction model should adequately predict the global orientation of a single kinematic chain.

Whereas the human datasets pose the challenge of having to model multiple kinematic chains simultaneously, the fish and mouse datasets raise different issues.
\begin{itemize}
    \item The datasets are smaller compared to H3.6m and CMU MoCap.
    \item The fish dataset consists of a long kinematic chain of 21 joints. There are 4 action types involved for the fish experiments, namely 1) `Scooting': straight line movement, 2) `J-turn': slight curvature (30 to 60 deg) forming a characteristic J shape  3) `C-turn': body curves into a C shape as the fish turns backwards and 4) `Routine-turn': a routine angular turn larger than 60 deg. The differences between these actions are quite subtle.
    \item Lack of activity classification for mouse dataset. Animal motions are generally more rapid paced and prone to environmental stimuli, thus displaying a higher degree of randomness than human motion. Consequently, for longer term forecasting, a more reasonable performance measure would be the plausibility and naturalness of the generated motion.
\end{itemize}

We benchmarked our \emph{AHMR} network against the ERD~\cite{erd} and ResGRU~\cite{residualgru}. The complex architectures and specific design elements in other methods make them highly intractable and difficult to adapt to other datasets. For example, large modifications and preprocessing have to be done for the convolutional blocks in CEM~\cite{li2018convolutional} or the graph networks in Traj-GCN~\cite{mao2019learning} and DMGNN~\cite{li2020dynamic} in order for them to be functional on different datasets. Our \emph{AHMR} network, on the contrary, is highly amenable and generalizable to any skeletal motion dataset. It operates directly on the fish and mouse dataset and we only require fine-tuning of a single hyperparameter, the dimension of the hidden units.

\begin{table}[ht!]
\centering
\begin{tabular}{|l|c|c|c|c|c|c|}
\hline
Time (ms) & 80 & 160 & 320 & 400 & 720 & 1,000 \\ \hline
\multicolumn{7}{c}{Fish}\\ \cline{1-7}
ERD~\cite{erd} & 0.65 & 0.64 & 0.74 & 0.90 & 1.29 & 1.65 \\ \hline
Zero-Velocity & 0.77 & 0.87 & 0.94 & 0.92 & 0.93 & 0.97 \\ \hline
Res-GRU~\cite{residualgru} & 0.76 & 0.71 & 0.60 & 0.53 & 0.77 & 0.86 \\ \hline
AHMR (ours) & \textbf{0.43} & \textbf{0.45} & \textbf{0.46} & \textbf{0.38} & \textbf{0.45} & \textbf{0.42} \\ \hline
\multicolumn{7}{c}{Mouse} \\ \cline{1-7}
ERD~\cite{erd} & 0.87 & 0.92 & 1.07 & 1.29 & 1.58 & 1.91 \\ \hline
Zero-Velocity & 0.49 & 0.71 & 0.97 & 1.21 & 1.33 & \textbf{1.28} \\ \hline
Res-GRU~\cite{residualgru} & 0.67 & 0.88 & 0.99 & 1.11 & 1.47 & 1.75 \\ \hline
AHMR (ours) & \textbf{0.47} & \textbf{0.68} & \textbf{0.76} & \textbf{0.83} & \textbf{1.07} & \textbf{1.28}\\ \hline
\end{tabular}
\caption{\footnotesize{Performance evaluation (in MAE) of the comparison methods for the Fish and Mouse datasets of~\cite{Liex}.}}
\label{tab:fishmouse}
\end{table}

Table~\ref{tab:fishmouse} presents the performance comparison on animal datasets, where the best performance is highlighted in boldface. From the table, we can see that \emph{AHMR} consistently and significantly outperforms the state-of-the-art methods on both datasets.

It would be more instructive to look at the visual results. A sample forecasting result for the fish dataset is shown in Figure~\ref{fig:Fish}. The long kinematic chain in the fish skeletal anatomy resulted in modeling difficulties for the competing methods. For ERD~\cite{erd}, the predicted fish pose demonstrate severe distortions and a zigzagged contour. ResGRU~\cite{residualgru} on the other hand, predicts a wrong direction for the fish motion and suffers from the issue of quickly converging to a motionless state. In contrast, \emph{AHMR} retains streamlined shapes and the curvature of the predicted pose remains smooth and natural.

The highly stochastic and fast moving nature of the mouse led to difficulties in accurate forecasting. A sample forecasting sequence on the mouse dataset is displayed in Figure~\ref{fig:Mouse}. ERD~\cite{erd} predicted unnatural motion where the mouse curled up into a distorted pose. ResGRU~\cite{residualgru} predicted a sequence where the only motion was a global rotation of the mouse and did not demonstrate motion along the other joints. \emph{AHMR} obtains fairly accurate prediction with natural and plausible motion. These results reveal that the motion context modeling of \emph{AHMR} is more effective.

\section{Conclusion}
We have addressed the problem of motion prediction from three directions, 1) an optimized Stiefel manifold parameterization of the data input; 2) an attention-based hierarchical motion recurrent network, which can effectively model motion contexts at multiple scales; and 3) geometrically motivated loss functions including a geodesic loss and forward kinematics loss which gives a much better measure of the discrepancies between prediction and ground truth. These three elements are seamlessly integrated into the proposed framework which naturally preserves the skeletal articulation of the underlying objects, and consistently delivers smooth motions. Extensive results on human, fish and mouse datasets demonstrate the efficacy of our approach. Strengths and limitations of existing methods are studied, with interesting findings and insights presented. We have further discussed on the efficiency, loss function, and variants of the proposed method, as well as the contributions of key components. Future work includes investigation into multi-subject motion predictions, as well as conditional motion synthesis.

\section*{Acknowledgment}
This paper is partly supported by the National Key R\&D Program of China (Grant No. 2018YFB1404102) and the Key R\&D Program of Zhejiang Province (No. 2021C01104), the A*STAR Computing \& Information Science Scholarship and the MOE Tier-1 project (Grant No: RG94/20) in Singapore, and the NSERC Discovery Grant and the UAHJIC grants in Canada.

\ifCLASSOPTIONcaptionsoff
 \newpage
\fi

\bibliographystyle{IEEEtran}
\bibliography{References}

\begin{IEEEbiography}[{\includegraphics[width=1in,height=1.3in,clip]{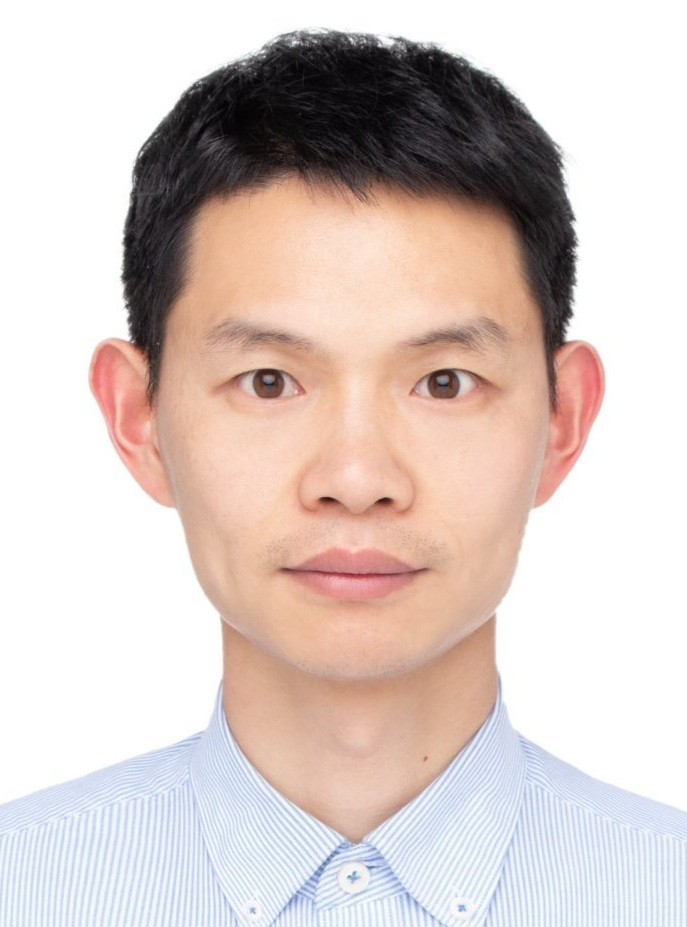}}]{Zhenguang Liu} was a Research Fellow with the National University of Singapore and the Singapore Agency for Science, Technology, and Research (A* STAR). He received the B.E. degree from Shandong University and the Ph.D. degree from Zhejiang University, China, in 2010 and 2015, respectively. He is currently a Professor of computer science with the School of Computer and Information Engineering, Zhejiang Gongshang University. Various parts of his work have been published in top-tier venues including CVPR, ICCV, TKDE, AAAI, ACM MM, INFOCOM, IJCAI, WWW, TMC, WWW. Dr. Liu has served as technical program committee member for top-tier conferences such as ACM MM, CVPR, AAAI, IJCAI, and ICCV, session chair of ICGIP, local chair of KSEM, and reviewer for top-tier journals IEEE TVCG, IEEE TPDS, IEEE TMM, ACM TOMM, etc. His research interests include motion and tracking, multimedia data analysis, and blockchain security.
\end{IEEEbiography}

\begin{IEEEbiography}[{\includegraphics[width=1in,height=1.3in,clip]{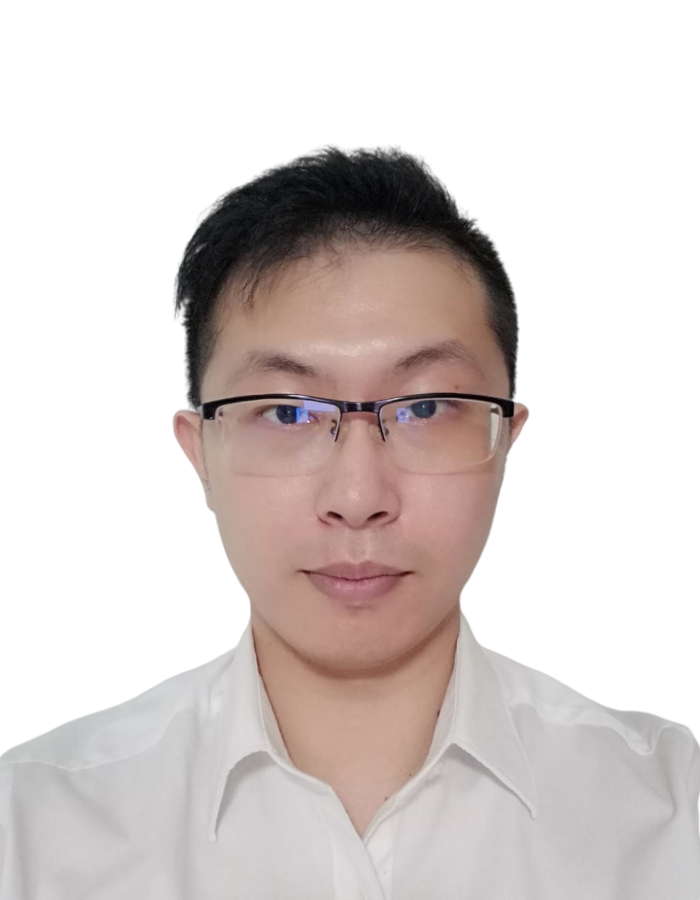}}]{Shuang Wu} is a PhD candidate with the School of Computer Science and Engineering at the Nanyang Technological University. He obtained a MSc and BSc from the National University of Singapore as well as a Dipl\^{o}me d'Ing\'{e}nieur from the \'{E}cole Polytechnique, specializing in particle physics. Prior to starting his PhD, he has worked at A*STAR, Singapore and CERN, Switzerland. His research interests include deep learning, computer vision and optimal transport.
\end{IEEEbiography}

\vfill

\begin{IEEEbiography}[{\includegraphics[width=1in,height=1.3in,clip]{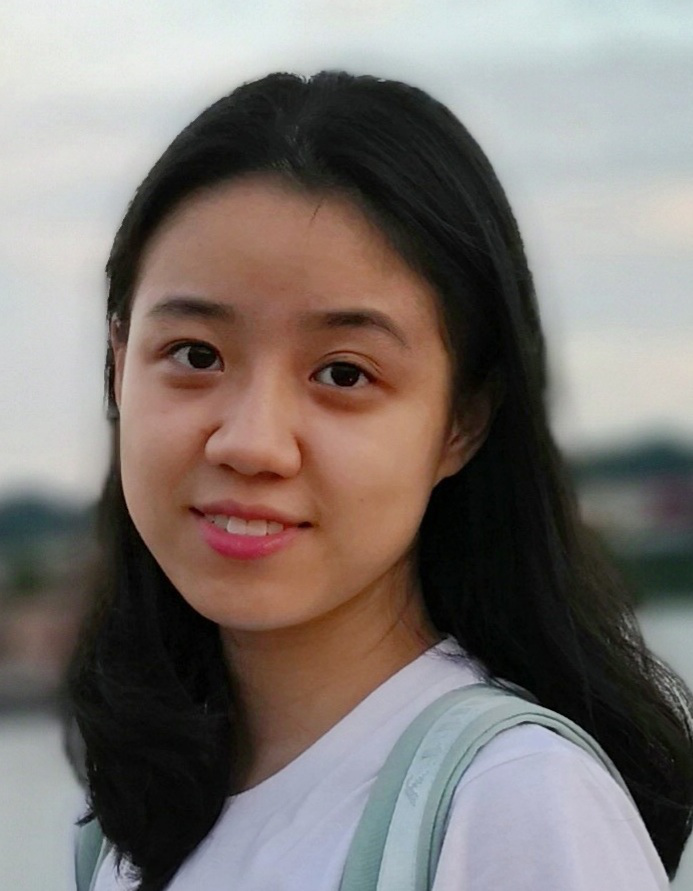}}]{Shuyuan Jin} 
is a software engineer at Facebook. Before joining Facebook, she graduated from the National University of Singapore in 2020. She had internship experience at A*STAR, Autodesk, Goldman Sachs, Quantedge hedge fund, and NUS Adaptive Computing Lab. Her main research focus is on deep learning and robotic planning.
\end{IEEEbiography}

\begin{IEEEbiography}[{\includegraphics[width=1in,height=1.3in,clip]{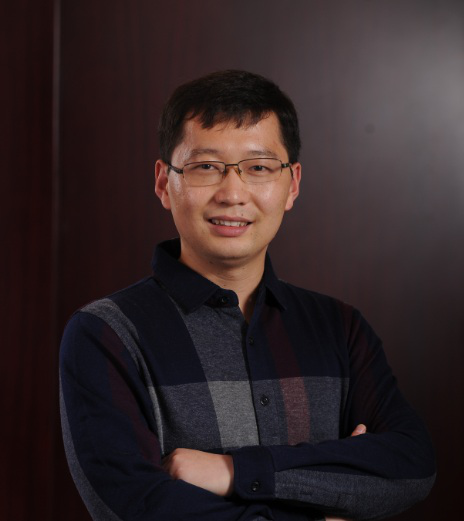}}]{Shouling Ji} 
is a Professor in the College of Computer Science and Technology at Zhejiang University and a Research Faculty in the School of Electrical and Computer Engineering at Georgia Institute of Technology. He received a Ph.D. degree in Electrical and Computer Engineering from Georgia Institute of Technology and a Ph.D. degree in Computer Science from Georgia State University. His research interests include AI Security, Data-driven Security, Software and System Security, and Big Data Analytics.
\end{IEEEbiography}

\begin{IEEEbiography}[{\includegraphics[width=1in,height=1.3in,clip]{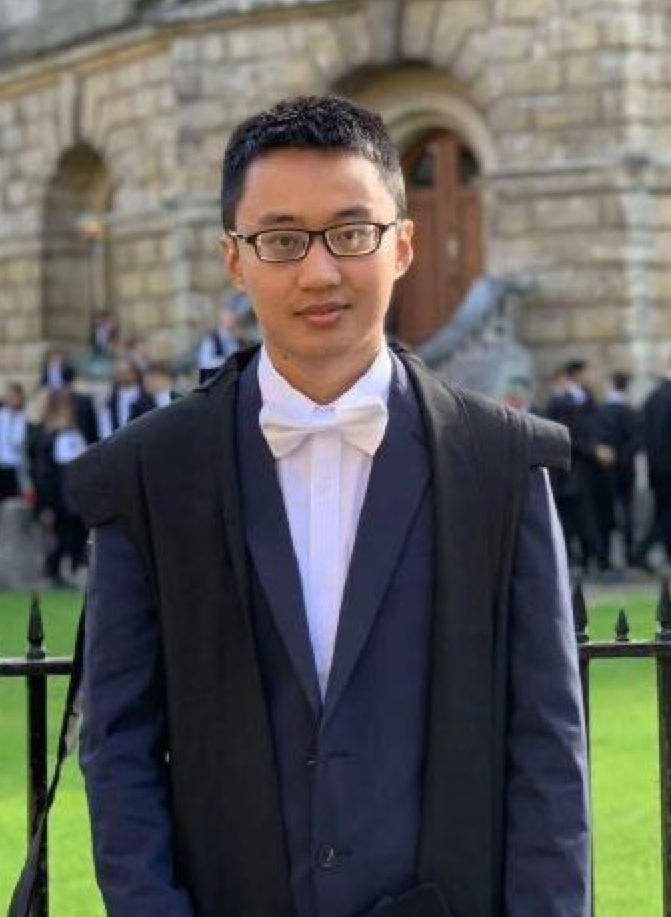}}]{Qi Liu}
is a DPhil student at the University of Oxford, advised by Matt Kusner and Phil Blunsom. Previously, he worked at Facebook AI Research in New York under the supervision of Douwe Kiela. He obtained his M.Sc. degree from National University of Singapore in 2016, working with Anthony K.H. Tung, and his B.Eng degree from Shandong University in 2014.
\end{IEEEbiography}

\begin{IEEEbiography}[{\includegraphics[width=1in,height=1.3in,clip]{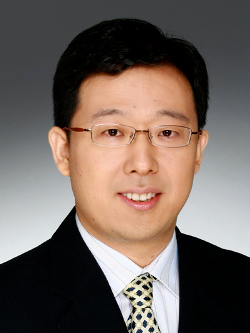}}]{Shijian Lu}
is an Assistant Professor with the School of Computer Science and Engineering at the Nanyang Technological University, Singapore. He received his PhD in electrical and computer engineering from the National University of Singapore. His major research interests include image and video analytics, visual intelligence, and machine learning.
\end{IEEEbiography}

\begin{IEEEbiography}[{\includegraphics[width=1in,height=1.3in,clip]{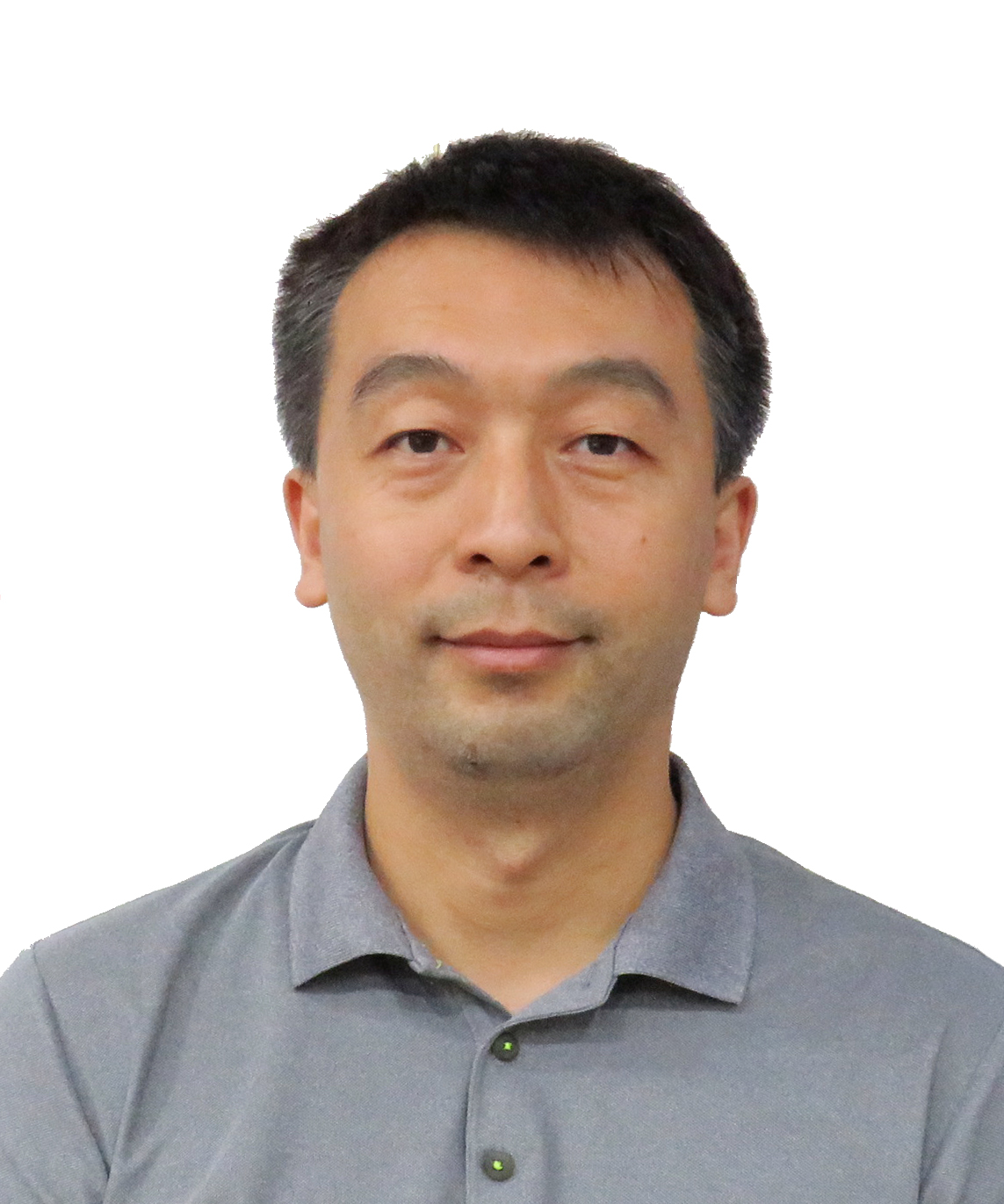}}]{Li Cheng} 
is an Associate Professor with the Department of Electrical and Computer Engineering, University of Alberta where he received his PhD in Computing Science. He is a senior member of IEEE. Prior to joining University of Alberta in year 2018, he worked at A*STAR, Singapore, TTI-Chicago, USA, and NICTA, Australia. His research expertise is mainly on computer vision and machine learning.
\end{IEEEbiography}

\vfill

\end{document}